\documentclass[sigconf]{acmart}

\usepackage{arydshln} 
\usepackage{booktabs}

\usepackage{graphicx}
\usepackage{epsfig}
\usepackage{algorithm}
\usepackage{algorithmic}
\usepackage{booktabs}
\usepackage{mathrsfs}
\usepackage{color}
\usepackage{textcomp}
\usepackage{bbding}
\usepackage{pifont}
\usepackage{wasysym}
\usepackage{subcaption}
\usepackage{xcolor,colortbl}
\usepackage{overpic}

\usepackage{microtype}
\usepackage{multirow}
\usepackage{makecell}
\usepackage{hhline}
\usepackage[american]{babel}
\usepackage{cuted}
\usepackage{ragged2e}
\usepackage{lipsum}

\definecolor{lightgray}{gray}{.94}
\definecolor{tinygray}{gray}{.96}

\usepackage{multirow}
\usepackage{epsfig}
\usepackage{adjustbox}
\usepackage[pagebackref=true,
            breaklinks=true,colorlinks,
            citecolor=blue,linkcolor=red,
            bookmarks=false]{}
\AtBeginDocument{%
  \providecommand\BibTeX{{%
    \normalfont B\kern-0.5em{\scshape i\kern-0.25em b}\kern-0.8em\TeX}}}

\copyrightyear{2024}
\acmYear{2024}
\setcopyright{acmlicensed}
\acmConference[MM '24] {Proceedings of the 32nd ACM International Conference on Multimedia}{October 28--November 1, 2024}{Melbourne, VIC, Australia.}
\acmBooktitle{Proceedings of the 32nd ACM International Conference on Multimedia (MM '24), October 28--November 1, 2024, Melbourne, VIC, Australia}
\acmISBN{979-8-4007-0686-8/24/10}
\acmDOI{10.1145/3664647.3681088}

\acmISBN{979-8-4007-0686-8/24/10}

\begin{document}
\begin{sloppypar}
\title{Efficient Face Super-Resolution via Wavelet-based Feature Enhancement Network}


% \author{Wenjie Li$^{1}$, Heng Guo$^{1,*}$, Xuannan Liu$^{1}$, Kongming Liang$^{1}$, Jiani Hu$^{1}$, Zhanyu Ma$^{1}$, Jun Guo$^{1}$}
%         % \authornote{Corresponding author}
%         \affiliation{
%          \institution{
%             $^1$ Beijing University of Posts and Telecommunications, Beijing, China
%         }
%         \city{}
%         \country{}
%         }
%         \email{{cswjli, guoheng, liuxuannan, liangkongming, hujiani, mazhanyu, guojun}@bupt.edu.cn}
		% \renewcommand{\shortauthors}{Wenjie Li et al.}

\author{Wenjie Li}
\affiliation{%
  \institution{Beijing University of Posts and Telecommunications}
  % \streetaddress{1 Th{\o}rv{\"a}ld Circle}
  \city{Beijing}
  \country{China}}
\email{cswjli@bupt.edu.cn}

\author{Heng Guo}
\authornote{Corresponding author}
\affiliation{%
  \institution{Beijing University of Posts and Telecommunications}
  \city{Beijing}
  \country{China}}
\email{guoheng@bupt.edu.cn}

\author{Xuannan Liu}
\affiliation{%
  \institution{Beijing University of Posts and Telecommunications}
  \city{Beijing}
  \country{China}}
\email{liuxuannan@bupt.edu.cn}

\author{Kongming Liang}
\affiliation{%
  \institution{Beijing University of Posts and Telecommunications}
  \city{Beijing}
  \country{China}}
\email{liangkongming@bupt.edu.cn}

\author{Jiani Hu}
\affiliation{%
  \institution{Beijing University of Posts and Telecommunications}
  \city{Beijing}
  \country{China}}
\email{jnhu@bupt.edu.cn}

\author{Zhanyu Ma}
\affiliation{%
  \institution{Beijing University of Posts and Telecommunications}
  \city{Beijing}
  \country{China}}
\email{mazhanyu@bupt.edu.cn}

\author{Jun Guo}
\affiliation{%
  \institution{Beijing University of Posts and Telecommunications}
  \city{Beijing}
  \country{China}}
\email{guojun@bupt.edu.cn}

\renewcommand{\shortauthors}{Wenjie Li et al.}

\begin{abstract} 
    Face super-resolution aims to reconstruct a high-resolution face image from a low-resolution face image. Previous methods typically employ an encoder-decoder structure to extract facial structural features, where the direct downsampling inevitably introduces distortions, especially to high-frequency features such as edges. To address this issue, we propose a wavelet-based feature enhancement network, which mitigates feature distortion by losslessly decomposing the input feature into high and low-frequency components using the wavelet transform and processing them separately. To improve the efficiency of facial feature extraction, a full domain Transformer is further proposed to enhance local, regional, and global facial features. Such designs allow our method to perform better without stacking many modules as previous methods did. Experiments show that our method effectively balances performance, model size, and speed. Code link: \url{https://github.com/PRIS-CV/WFEN}. 
\end{abstract}


%    <concept>
%        <concept_id>10010147.10010178.10010224.10010245.10010254</concept_id>
%        <concept_desc>Computing methodologies~Reconstruction</concept_desc>
%        <concept_significance>500</concept_significance>
%        </concept>
%  </ccs2012>
% \ccsdesc[500]{Computing methodologies~Reconstruction}






\maketitle

\begin{figure}[ht]
    \vspace{-2mm}
    \begin{tabular}{cc}\hspace{-4mm}
    \begin{picture}(120,100)
    \put(0,0){\includegraphics[width=0.24\textwidth]{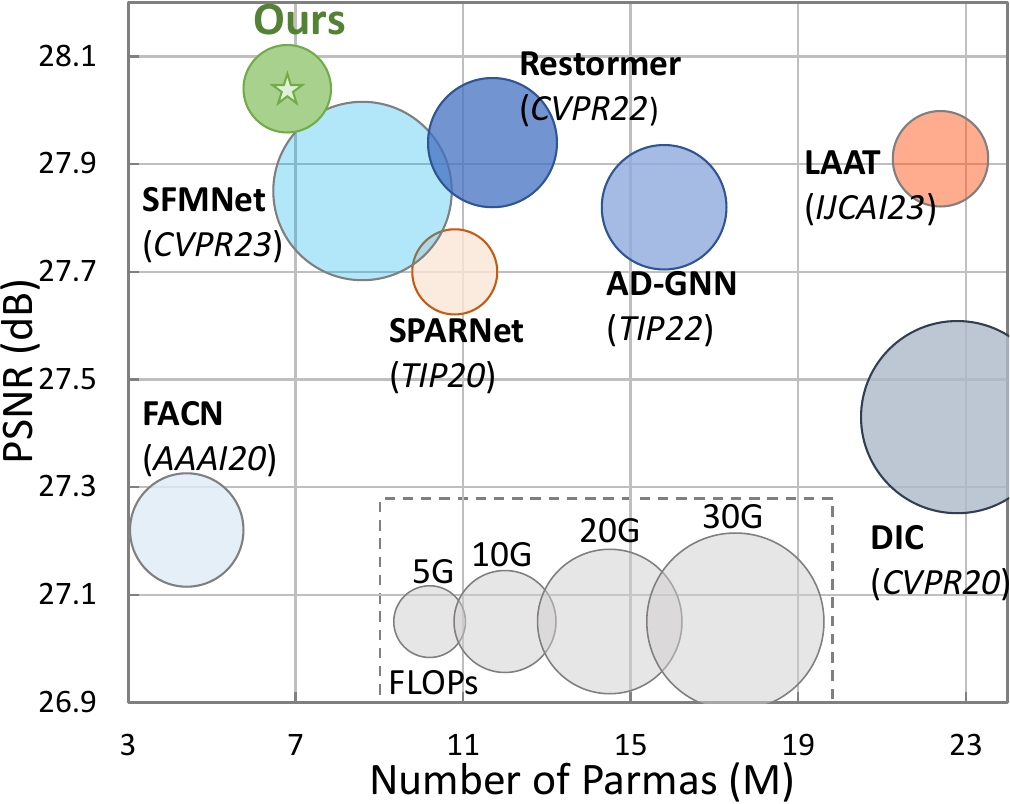}}
    \put(20,61.5){\fontsize{4.5pt}{1pt}\selectfont~\cite{wang2023spatial}}
    
    \put(26.5,45.5){\fontsize{4.5pt}{1pt}\selectfont~\cite{xin2020facial}}

    \put(59,50.5){\fontsize{4.5pt}{1pt}\selectfont~\cite{chen2020learning}}

    \put(88,61){\fontsize{4.5pt}{1pt}\selectfont~\cite{bao2022attention}}

    \put(81,87.5){\fontsize{4.5pt}{1pt}\selectfont~\cite{zamir2022restormer}}

    \put(110.5,30.5){\fontsize{4.5pt}{1pt}\selectfont~\cite{ma2020deep}}

    \put(100,66){\fontsize{4.5pt}{1pt}\selectfont~\cite{LAATransformer}}
    \end{picture}

    \begin{picture}(120,100)
    \put(0,0){\includegraphics[width=0.24\textwidth]{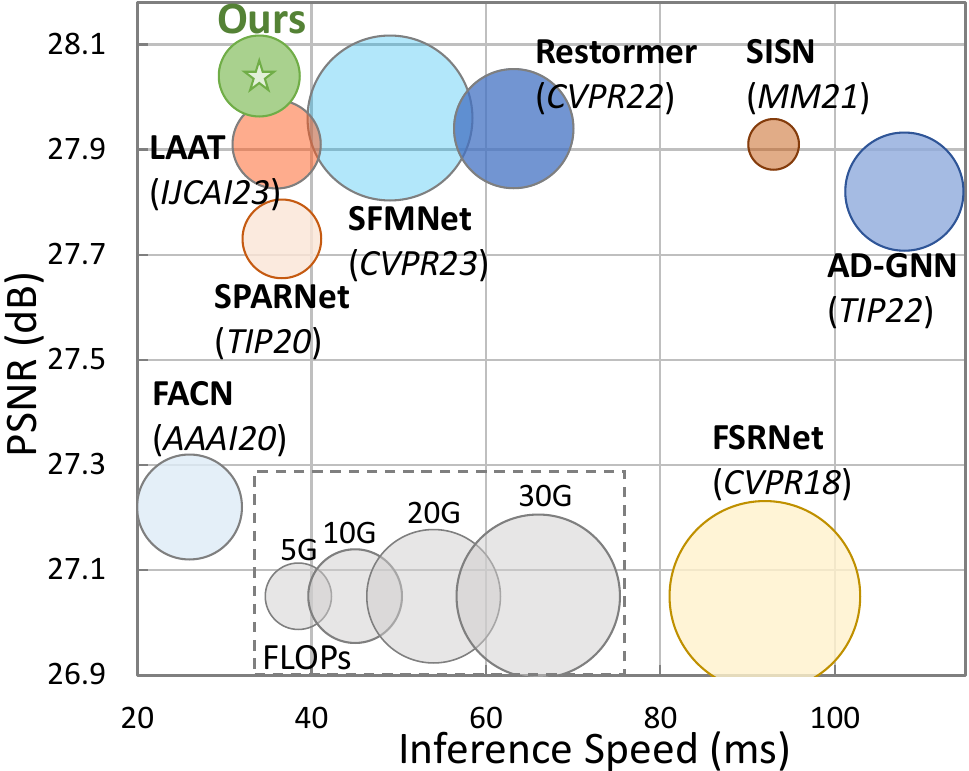}}
    \put(58,68){\fontsize{4.5pt}{1pt}\selectfont~\cite{wang2023spatial}}

    \put(28.5,46){\fontsize{4.5pt}{1pt}\selectfont~\cite{xin2020facial}}

    \put(39.5,52.5){\fontsize{4.5pt}{1pt}\selectfont~\cite{chen2020learning}}

    \put(72,79){\fontsize{4.5pt}{1pt}\selectfont~\cite{zamir2022restormer}}

    \put(20,66.8){\fontsize{4.5pt}{1pt}\selectfont~\cite{LAATransformer}}

    \put(101.5,89){\fontsize{4.5pt}{1pt}\selectfont~\cite{lu2021face}}

    \put(107,52){\fontsize{4.5pt}{1pt}\selectfont~\cite{bao2022attention}}

    \put(102.5,40.5){\fontsize{4.5pt}{1pt}\selectfont~\cite{chen2018fsrnet}}

    \end{picture}
    
    \\
    & \hspace{-88mm} 

    \fontsize{7pt}{1pt}\selectfont (a) PSNR, FLOPs and Params Tradeoffs. 
    
    \hspace{2mm} 
    
    \fontsize{7pt}{1pt}\selectfont (b) PSNR, FLOPs and Speed Tradeoffs.

    \\

    \end{tabular}
    \vspace{-2mm}
 \caption{Efficiency trade-offs between ours and existing methods on CelebA~\cite{liu2015deep} test set. Our method achieves a balance in terms of PSNR, model size, and speed.}
 \vspace{-2mm}
\label{Fig:balance}
\end{figure}

\section{Introduction}
Face super-resolution (FSR), also known as face hallucination, aims to convert a low-resolution (LR) face image into a high-resolution (HR) face image. Different from image super-resolution, FSR focuses on reconstructing essential structural information about the face, including facial contours and the shape of facial components. This paper aims to propose a high-fidelity FSR method while maintaining efficiency in model size and inference speed, as depicted in Fig.~\ref{Fig:balance}.

% Encoder-decoder structure facilitates the model to grasp the overall structure of the face during the encoder stage with a larger receptive field and enhances facial details during the decoder stage. Specifically, the encoder initially downsamples the LR input, extracting facial features at various scales. Subsequently, the decoder progressively upsamples the features from the encoder output, refines details, and ultimately generates the HR output. Hence, existing FSR methods~\cite{chen2020learning,bao2022attention,LAATransformer} typically apply an encoder-decoder structure.

\begin{figure}
\captionsetup{font=small}
\scriptsize
\hspace{-0.2cm}
\scalebox{1}{
\begin{tabular}{c@{\extracolsep{0em}}c@{\extracolsep{0.03em}}c@{\extracolsep{0.03em}}c@{\extracolsep{0.03em}}c@{\extracolsep{0.03em}}c@{\extracolsep{0.03em}}}
\includegraphics[width=0.089\textwidth]{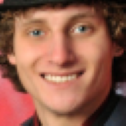}~~
&\includegraphics[width=0.089\textwidth]{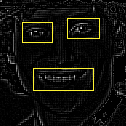}~~
&\includegraphics[width=0.089\textwidth]{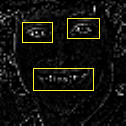}~~
&\includegraphics[width=0.089\textwidth]{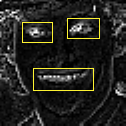}~~
&\includegraphics[width=0.089\textwidth]{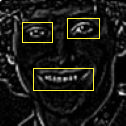}~~\\

\includegraphics[width=0.089\textwidth]{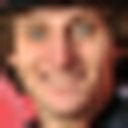}~~
&\includegraphics[width=0.089\textwidth]{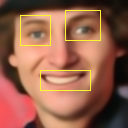}~~
&\includegraphics[width=0.089\textwidth]{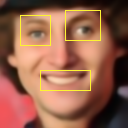}~~
&\includegraphics[width=0.089\textwidth]{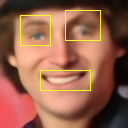}~~
&\includegraphics[width=0.089\textwidth]{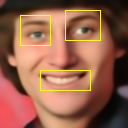}~~\\
\scriptsize (a) Input \footnotemark[1] & \hspace{-3mm}
\scriptsize (b) Bicubic & \hspace{-2mm}
\scriptsize (c) Stride Conv  & \hspace{-3mm}
\scriptsize (d) AvgPool & \hspace{-2mm}
\scriptsize (e) Ours & \hspace{-3mm}\\
\end{tabular}}
\vspace{-0.1cm}
\caption{Feature maps (first line) and FSR results (second line) with various downsampling methods: bicubic interpolation, stride convolution, average pooling, and our wavelet feature downsample. The loss of high-frequency features is pronounced in (a) and (b), while frequency-domain feature aliasing appears in (c). Ours is effective in preventing feature loss or frequency-domain aliasing.}
\label{fig:motivation}
\vspace{-0.1cm}
\end{figure}

Existing FSR methods~\cite{chen2020learning,bao2022attention,LAATransformer} typically apply an encoder-decoder structure, because it facilitates models to grasp the overall facial structure during the encoder stage with a larger receptive field and enhances facial details during the decoder stage. Specifically, the encoder initially downsamples LR inputs, extracting features at various scales. Subsequently, the decoder progressively upsamples features from encoder outputs, refines details, and ultimately generates HR outputs. However, previous methods neglect the impact of the chosen downsampling technique on FSR results. For example, some methods~\cite{chen2020learning,bao2022attention} employ bicubic interpolation or strided convolution for downsampling, which reduces the number of image pixels, potentially leading to the loss of facial details essential for FSR. As depicted in Fig.~\ref{fig:motivation} (b) and (c), bicubic interpolation and strided convolution lead to a significant loss of texture in the overall facial structure, resulting in distortions in the reconstruction of the face profile. Another example from ~\cite{kim2021edge} utilizes downsampling through avgpooling. As shown in Fig.~\ref{fig:motivation} (d), this results in the aliasing artifact of high and low-frequency facial features. This phenomenon is particularly evident in the eye features and significantly hampers the accurate representation of facial components.

To address the above problem, we propose to utilize discrete wavelet transform~\cite{mallat1989theory} (DWT) to enhance facial features. Specifically, following the Nyquist sampling theorem, the standard downsampling process involves a low-pass filter followed by downsampling. DWT can simulate standard downsampling by decomposing the input image \hbox{$\boldsymbol{I}$~$\in$~$\mathbb{R}^{H\times W }$} into four components at different frequencies, which consist of one low-frequency component \hbox{${\boldsymbol{I}_{LL}}$~$\in$~$\mathbb{R}^{\frac{H}{2} \times \frac{W}{2} }$} and three high-frequency components \hbox{$\{ {\boldsymbol{I}_{LH}},{\boldsymbol{I}_{HL}},{\boldsymbol{I}_{HH}}\} $~$\in$~$\mathbb{R}^{\frac{H}{2} \times \frac{W}{2} }$}. The low-frequency component ${\boldsymbol{I}_{LL}}$ can be approximated as the result obtained after low-pass filtering followed by downsampling. Simultaneously, the acquired high-frequency components can still be fused with low-frequency features to enhance features, such as face edges. As shown in Fig.~\ref{fig:motivation} (e), employing wavelet feature decomposition and fusion results in clearer facial contours, and there is no occurrence of frequency domain aliasing. This result shows that the use of DWT to decompose and fuse the high and low-frequency features can accomplish downsampling while ensuring a clear facial profile. Inspired by this observation, we introduce wavelet feature downsample (WFD) and wavelet feature upgrade (WFU). WFD aims to minimize distortion of crucial facial structures during downsampling in encoder phases. WFU aims to enhance facial contour by utilizing high-frequency features obtained through DWT in decoder phases.

% , to alleviate the problem of facial structure loss caused by downsampling    to prevent frequency domain aliasing

To better enhance the low-frequency facial information obtained after wavelet transform decomposition, we introduce a full-domain Transformer (FDT). Specifically, since the low-frequency information includes numerous detailed features of images, extracting comprehensive low-frequency information is crucial. Despite the Transformer demonstrating efficacy in handling low-frequency information, the Transformer utilized in existing FSR methods struggles to effectively concentrate on local (\emph{e.g.}, skin details), regional (\emph{e.g.}, components like eyes, noses), and global features (\emph{e.g.}, overall face profile). To address this issue, FDT is proposed to explore diverse receptive fields and uncover deeper correlations within facial features to extract more comprehensive information to enhance FSR.

\footnotetext[1]{The clearer input for the first line is to make the feature map easily observable and the contrast pronounced.}

By utilizing WFD and WFU to alleviate facial feature distortion and employing FDT for comprehensive extraction of facial features, our wavelet-based feature enhancement network (WFEN) achieves robust performance without the need for excessive network module stacking like previous methods. Results as Fig.~\ref{Fig:balance}, our WFEN demonstrates outstanding efficiency compared to state-of-the-art methods. In summary, the contributions of this paper are as follows:

\begin{itemize}
  \item We propose WFD and WFU modules utilizing wavelet transform to minimize the distortion of facial features and enhance face contour in the encoder-decoder structure.
  \item We propose an FDT module that extends interactions to the local, regional, and global levels, providing our model with richer facial receptive field information.
  \item We propose a WFEN, which is more efficient than state-of-the-art methods and achieves a better balance in performance, model size, and inference speed.
\end{itemize}

\section{Related Work}
Since our method enhances FSR performance through the application of wavelet theory, we present recent advances in FSR and discuss the utilization of wavelet transform in super-resolution. The difference with the main related methods can be seen in Table~\ref{tab:differnet}. 

\begin{table}[t]
\tiny
\setlength\tabcolsep{1pt}
\centering
\vspace{0mm}
    \caption{Comparison of encoder-decoder-based network design in existing methods.}
\vspace{-2mm}
\label{tab:differnet}
\resizebox{0.48\textwidth}{!}{
\begin{tabular}{l||c|l}
\toprule
\rowcolor{lightgray}
    \rowcolor{lightgray}
    \multicolumn{1}{l||}{\multirow{-1}{*}{Methods}} 
    & \multicolumn{1}{c|}{\multirow{-1}{*}{Wavelet-based}}
    & \multicolumn{1}{l}{\multirow{-1}{*}{Methods Features}}

    \\ 
    \hline\hline

    SPARNet~\cite{chen2020learning}  & No & Spatial attention\\ 
    Restormer~\cite{zamir2022restormer}  & No & Channel-based self-attention\\ 
    LAAT~\cite{LAATransformer}  & Yes & Feature fusion from coarse to fine\\ 
    SFMNet~\cite{wang2023spatial}  & No & Utilizing Fourier domain feature\\
    \textbf{Ours}  & Yes &  Mitigating feature corruption in downsample\\

    \bottomrule
\end{tabular}}
\end{table}
\hspace{-0mm}

\begin{figure*}
\begin{center}
\includegraphics[width=0.99\linewidth]{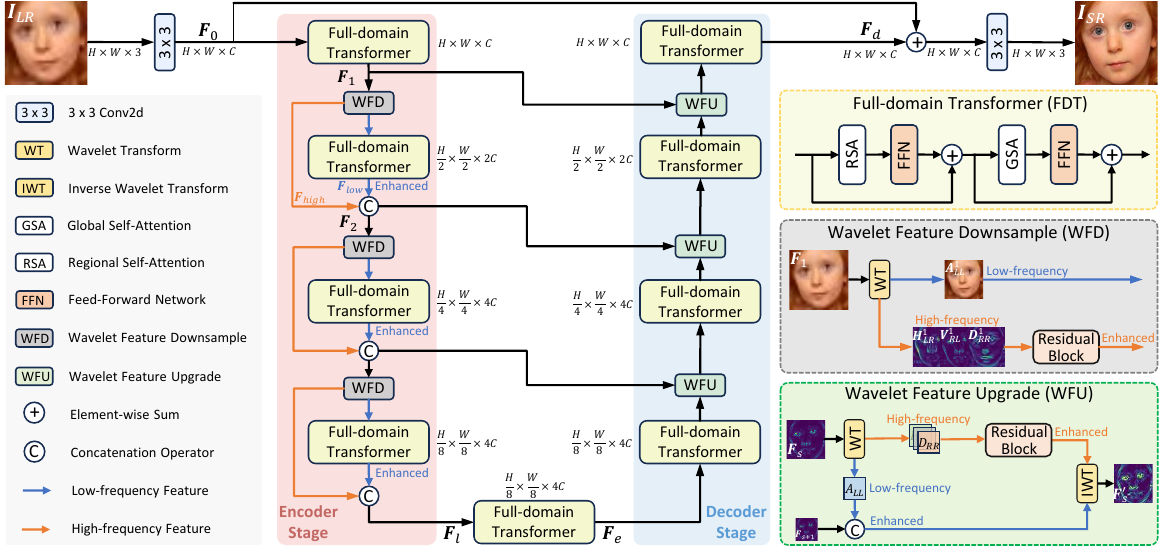}
\vspace{-0mm}
\end{center}
   \caption{Overview of our method, where the cascaded of WFD and WFU constitute the wavelet-based encoder-decoder structure.}
\label{fig:main}
\vspace{-0mm}
\end{figure*}

\subsection{Face Super-resolution}
Recently, numerous neural networks~\cite{li2023survey} for FSR have been proposed to enhance performance. Due to the highly structured nature of the human face, one category of methods aims to leverage facial priors, such as facial landmarks~\cite{ma2020deep}, facial parsing maps~\cite{chen2018fsrnet}, facial attribute~\cite{xin2020facial}, 3D facial shapes~\cite{hu2021face}, \emph{etc.}, to assist in the restoration process. However, incorporating the face prior estimation module will unavoidably introduce an additional computational burden. Moreover, accurately estimating facial geometric priors becomes highly challenging when dealing with very low-resolution face images, which results in a substantial distortion of FSR results.

Consequently, attention-based FSR methods have gained prominence. RAAN~\cite{xin2019residual} utilizes channel attention to extract face shape features. SPARNet~\cite{chen2020learning} introduces spatial attention to capture facial structural features. SISN~\cite{lu2021face} separately explores facial structural and texture details through the external-internal separation of attention. AD-GNN~\cite{bao2022attention} utilizes a series of spatial attention to explore feature relationships. To simulate long-distance modeling, FaceFormer~\cite{wang2022faceformer} leverages Transformer to capture global facial information. LAAT~\cite{LAATransformer} further enhances fine-grained regions by introducing a self-refinement mechanism into the Transformer. SCTANet~\cite{bao2023sctanet} integrates spatial attention and self-attention to leverage local and global features. SFMNet~\cite{wang2023spatial} employs frequency branching and spatial branching to extract global and local features, respectively. However, none of them consider the detrimental effect of downsampling in the encoder-decoder. Moreover, unlike these methods that focus only on local or global facial features, our method simultaneously focuses on local, regional, and global facial features.

\subsection{Wavelet Transform-based Methods}
Recently, wavelet theory has gained prominence in super-resolution. DWSR~\cite{guo2017deep} employs CNN representations on low-resolution wavelet subbands to recover missing details. Wavelet-SRNet~\cite{huang2017wavelet} reconstructs a face image from a sequence of wavelet coefficients of the HR corresponding to the LR learned by the network. SRCliqueNet~\cite{zhong2018joint} explores relationships between wavelet transform subbands to aid the reconstruction process. WaveletSRGAN~\cite{huang2019wavelet} learns wavelet information for predicting HR face images. WDRN~\cite{xin2020wavelet} utilizes a series of wavelet coefficients to reconstruct the HR image. JWSGN~\cite{zou2022joint} employs wavelet transform to reconstruct the frequency domain details of images. WTRN~\cite{li2022wavelet} reconstructs the texture by computing the correlation of the wavelet-transformed subbands with the reference image. LAAT~\cite{LAATransformer} and MRLPFNet~\cite{dong2023multi} employ a wavelet fusion module to combine shallow structures and deep details to recover realistic images in the frequency domain. Unlike these methods, we focus on utilizing wavelet transform to decompose the high and low-frequency components for lossless downsampling, which in turn reduces the feature corruption of downsampling in encoder-decoder structure.

\section{Proposed Method}
\subsection{Overview}
As shown in Fig.~\ref{fig:main}, from a given degraded face image \hbox{${{\boldsymbol{I}_{LR}}}$~$\in$~$\mathbb{R}^{H\times W \times 3}$}, we aim to reconstruct a clean face \hbox{${\boldsymbol{I}_{SR}}$~$\in$~$\mathbb{R}^{H\times W \times 3}$} by employing a wavelet-based encoder-decoder structure integrated with residual block and our full-domain transformer. The wavelet-based encoder-decoder structure encompasses our wavelet feature downsample in the encoder stage and wavelet feature upgrade in the decoder stage for downsampling and upsampling.

\begin{figure*}
\begin{center}
\includegraphics[width=0.99\linewidth]{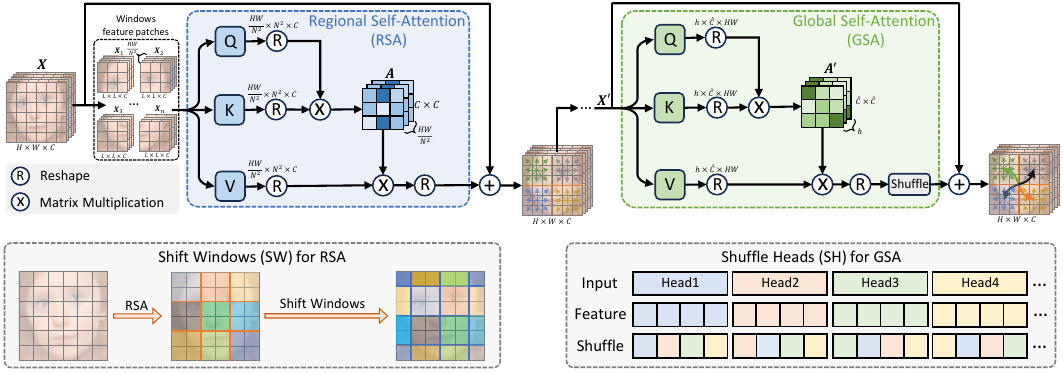}
\vspace{-0mm}
\end{center}
   \caption{Architecture of our full-domain Transformer, which can handle local, regional, and global facial features.}
\label{fig:attention}
\vspace{-0mm}
\end{figure*}

Specifically, our method initially extracts shallow facial features \hbox{${\boldsymbol{F}_{\rm{0}}}$~$\in$~$\mathbb{R}^{H\times W \times C}$} from \hbox{${{\boldsymbol{I}_{LR}}}$}, where \hbox{$H \times W$} denotes the spatial resolution, and \hbox{$C$} denotes the number of channels. Subsequently, \hbox{${\boldsymbol{F}_{\rm{0}}}$} undergoes hierarchical level-by-level processing through wavelet feature downsample, gradually transforming the \hbox{${\boldsymbol{F}_{\rm{0}}}$} into a low-resolution latent representation \hbox{${\boldsymbol{F}_{\rm{l}}}$~$\in$~$\mathbb{R}^{\frac{H}{8} \times \frac{W}{8} \times 4C}$}. At each level, the low-frequency part of the transform is fed to our full-domain Transformer, while the high-frequency part is fed to the residual block. During the bottleneck stage, situated between the encoder and decoder stages, a sequence of full-domain Transformers is employed to refine \hbox{${\boldsymbol{F}_{\rm{l}}}$} to obtain \hbox{${\boldsymbol{F}_{\rm{e}}}$~$\in$~$\mathbb{R}^{\frac{H}{8} \times \frac{W}{8} \times 4C}$}. Then, we incorporate wavelet feature upgrade before each decoding level, which effectively performs cross-scale feature fusion to obtain accurate depth features \hbox{${\boldsymbol{F}_{\rm{d}}}$~$\in$~$\mathbb{R}^{H\times W \times C}$}. Finally, the output \hbox{${\boldsymbol{F}_{\rm{d}}}$} from the decoder stage recovers a clean face image \hbox{${\boldsymbol{I}_{SR}}$} after residual concatenation and dimensionality reduction. In the following subsections, we provide a detailed description of the core modules we have constructed.

\subsection{Wavelet-based Encoder-Decoder Structure}
As depicted in Fig.~\ref{fig:main}, the central components of the wavelet-based encoder-decoder structure consist of a series of wavelet feature downsamples in the encoder and a series of wavelet feature upgrades in the decoder. They are tasked with progressively downsampling and upsampling, forming the main structure of our network. 

\paragraph{Wavelet Feature Downsample (WFD)} During the encoder process, downsampling is typically employed to decrease the size of the feature map. However, as mentioned above, existing methods overlook the irreversible distortion caused by downsampling, resulting in unclear edges in the FSR results. It occurs because traditional downsampling operations, which decrease resolution by merging neighboring pixels, can result in feature distortion, particularly in regions with significant gradient changes, due to the reduction in sampling points. In this context, as shown in Fig.~\ref{fig:main}, we introduce a WFD using wavelet transform to alleviate this phenomenon. Additional information about wavelet transform can be found in our \emph{supplementary materials within our code link}.

For input facial feature \hbox{${\boldsymbol{F}_{\rm{1}}}$~$\in$~$\mathbb{R}^{H\times W \times C}$}, we initially apply a wavelet transform $\mathcal{WT}$, allowing us to decompose \hbox{${\boldsymbol{F}_{\rm{1}}}$} into four sub-wavelet bands: low-pass feature \hbox{$\boldsymbol{A}_{LL}^1$}, and high-frequency facial in horizontal, vertical, and diagonal directions \hbox{$\boldsymbol{H}_{LR}^1$, $\boldsymbol{V}_{RL}^1$}, and \hbox{$\boldsymbol{D}_{RR}^1$}:
\begin{equation}
\{ \boldsymbol{A}_{LL}^1,\boldsymbol{H}_{LR}^1,\boldsymbol{V}_{RL}^1,\boldsymbol{D}_{RR}^1\}  = \mathcal{WT}(\boldsymbol{{F}_1}) ,
\end{equation}
where \hbox{$\{ \boldsymbol{A}_{LR}^1,\boldsymbol{H}_{LR}^1,\boldsymbol{V}_{RL}^1,\boldsymbol{D}_{RR}^1\}  $~$\in$~$\mathbb{R}^{\frac{H}{2}\times \frac{H}{2} \times C}$}. As the low-frequency part predominantly carries essential information in the image, we focus on low-frequency face details on the main path while paying attention to high-frequency face textures on the residual path: 
\begin{equation}
\boldsymbol{{F}}_{low},\boldsymbol{{F}}_{high} = \mathcal{T}\left( {\boldsymbol{A}_{LL}^1} \right),\mathcal{R}\left( {\boldsymbol{H}_{LR}^1,\boldsymbol{V}_{RL}^1,\boldsymbol{D}_{RR}^1} \right),
\end{equation}
where \hbox{${\boldsymbol{F}_{\rm{low}}}$} denotes the enhanced low-frequency features, \hbox{${\boldsymbol{F}_{\rm{high}}}$} denotes the enhanced high-frequency features, \hbox{$\mathcal{T}$} denotes our full-domain Transformer, and \hbox{$\mathcal{R}$} denotes residual block. We opt for different structures to extract high and low-frequency features because prior researches~\cite{si2022inception,li2023feature} indicate that Transformer is more sensitive to low-frequency features, while CNN is more sensitive to high-frequency features. Next, the full downsampled enhanced feature \hbox{${\boldsymbol{F}_{\rm{2}}}$~$\in$~$\mathbb{R}^{\frac{H}{2}\times \frac{H}{2} \times 2C}$} is obtained by fusing \hbox{${\boldsymbol{F}_{\rm{low}}}$} and \hbox{${\boldsymbol{F}_{\rm{high}}}$}. With this thoughtful design, our model can enhance efficiency in handling both high and low-frequency facial features.

\begin{table*}[t!]
\tiny
\setlength\tabcolsep{2.5pt}
\centering
\vspace{0mm}
    \caption{\fontsize{8.9pt}{1pt}\selectfont Quantatitive evaluation for $\times 8$ FSR on CelebA~\cite{liu2015deep} and Helen~\cite{le2012interactive} test sets. The best and second-best results are emphasized in \bf{bold} and underlined. Our method achieves the best results with the second least computational load and speed.}
\vspace{-0mm}
\label{tab:performance}
\resizebox{0.99\textwidth}{!}{
\begin{tabular}{l|l|l|l||ccccc|cccc}
\toprule
\rowcolor{lightgray}
& & & & \multicolumn{5}{c|}{CelebA~\cite{liu2015deep}} 
& \multicolumn{4}{c}{Helen~\cite{le2012interactive}} 
\\ 
\cmidrule{5-13}
    \rowcolor{lightgray}
    \multicolumn{1}{l|}{\multirow{-2}{*}{Methods}} 
    & \multicolumn{1}{c|}{\multirow{-2}{*}{Params}}
    & \multicolumn{1}{c|}{\multirow{-2}{*}{FLOPs}}
    & \multicolumn{1}{c||}{\multirow{-2}{*}{Speed}}
    & PSNR$\uparrow$  & SSIM$\uparrow$    & LPIPS$\downarrow$  
    & VIF$\uparrow$ & ID$\uparrow$  
    
    & PSNR$\uparrow$  & SSIM$\uparrow$    & LPIPS$\downarrow$  
    & VIF$\uparrow$   
    \\ 
    \hline\hline
    Bicubic     & -   & -  & -     & 23.61  & 0.6779  & 0.4899  & 0.1821  & 5.9$\%$  & 22.95  & 0.6762  & 0.4912 & 0.1745    \\ 
    FSRNet~\cite{chen2018fsrnet}     & 27.5M   & 40.7G  & 89.8ms     & 27.05  & 0.7714  & 0.2127  & 0.3852  & 66.7$\%$  & 25.45 & 0.7364  & 0.3090  & 0.3482    \\ 
    FACN~\cite{xin2020facial}     & \bf{4.4M}   & 12.5G  & \bf{26.7ms}     & 27.22  & 0.7802  & 0.2828  & 0.4366 & 67.1$\%$  & 25.06  & 0.7189  & 0.3113  & 0.3702   \\ 
    DIC~\cite{ma2020deep}     & 22.8M   & 35.5G  & 120.5ms     & 27.42  & 0.7840  & 0.2129  & 0.4234  & 71.6$\%$  & 26.15  & 0.7717  & \underline{0.2158} & 0.4085    \\ 
    SPARNet~\cite{chen2020learning}     & 10.6M   & \bf{7.1G}  & 36.6ms     & 27.73  & 0.7949  & 0.1995  & 0.4505  & 80.3$\%$  & 26.43  & 0.7839  & 0.2674 & 0.4262    \\ 
    % SISN~\cite{lu2021face}     & 9.8M   & 2.5G  & 93.3ms     & 27.91  & 0.7971  & 0.2005  & 0.4785  & -  & 26.64  & 0.7908  & 0.2571 & 0.4623  & -  \\
    AD-GNN~\cite{bao2022attention}     & 15.8M   & 15.0G  & 107.9ms     & 27.82  & 0.7962  & 0.1937  & 0.4470  & 81.2$\%$  & 26.57  & 0.7886  & 0.2432 & 0.4363    \\ 
    Restormer-M~\cite{zamir2022restormer}     & 11.7M   & 16.1G  & 63.2ms     & 27.94  & \underline{0.8027}  & 0.1933  & 0.4624  & 82.4$\%$  & \underline{26.91}  & \underline{0.8013}  & 0.2258 & \underline{0.4595}   \\ 
    LAAT~\cite{LAATransformer}     & 22.4M   & 8.9G  & 35.1ms     & 27.91  & 0.7994  & \underline{0.1879}  & 0.4624  & \underline{84.8$\%$}  & 26.89  & 0.8005  & 0.2255  & 0.4569   \\ 
    SFMNet~\cite{wang2023spatial}     & 8.6M   & 30.6G  & 49.2ms     & \underline{27.96}  & 0.7996  & 0.1937  & \underline{0.4644}  & 84.6$\%$  & 26.86  & 0.7987  & 0.2322  & 0.4573    \\ 
    \bf{Ours}     & \underline{6.8M}   & \underline{7.5G}  & \underline{33.9ms}     & \bf{28.04}  & \bf{0.8032}  & \bf{0.1803}  & \bf{0.4682}  & \bf{86.8$\%$}  & \bf{27.01}  & \bf{0.8051}  & \bf{0.2148} & \bf{0.4631}  \\ 
    
    \toprule
\end{tabular}}
\end{table*}
\hspace{-0mm}

\begin{figure*}[!htbp]
\captionsetup{font=small}
\scriptsize
\hspace{0cm}
\scalebox{0.99}{
\begin{tabular}{c@{\extracolsep{0em}}@{\extracolsep{0.05em}}c@{\extracolsep{0.05em}}c@{\extracolsep{0.05em}}c@{\extracolsep{0.05em}}c@{\extracolsep{0.05em}}c@{\extracolsep{0.05em}}c@{\extracolsep{0.05em}}c@{\extracolsep{0.05em}}c@{\extracolsep{0.00em}}c@{\extracolsep{0.00em}}}

\includegraphics[width=0.092\textwidth]{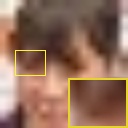}~~
		&\includegraphics[width=0.092\textwidth]{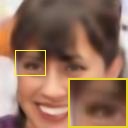}~~
		&\includegraphics[width=0.092\textwidth]{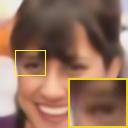}~~
		&\includegraphics[width=0.092\textwidth]{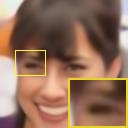}~~
        &\includegraphics[width=0.092\textwidth]{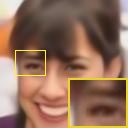}~~
		&\includegraphics[width=0.092\textwidth]{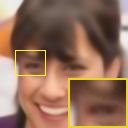}~~
		&\includegraphics[width=0.092\textwidth]{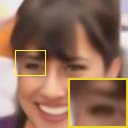}~~
        &\includegraphics[width=0.092\textwidth]{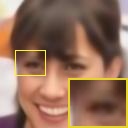}~~
        &\includegraphics[width=0.092\textwidth]{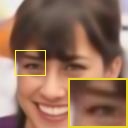}~~
        &\includegraphics[width=0.092\textwidth]{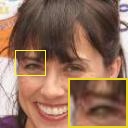}~~\\

\includegraphics[width=0.092\textwidth]{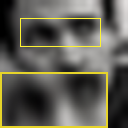}~~
		&\includegraphics[width=0.092\textwidth]{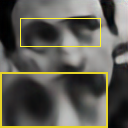}~~
		&\includegraphics[width=0.092\textwidth]{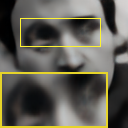}~~
		&\includegraphics[width=0.092\textwidth]{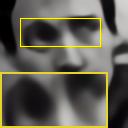}~~
        &\includegraphics[width=0.092\textwidth]{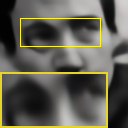}~~
		&\includegraphics[width=0.092\textwidth]{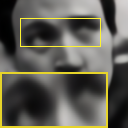}~~
		&\includegraphics[width=0.092\textwidth]{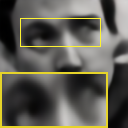}~~
        &\includegraphics[width=0.092\textwidth]{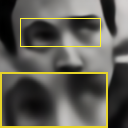}~~
        &\includegraphics[width=0.092\textwidth]{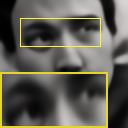}~~
        &\includegraphics[width=0.092\textwidth]{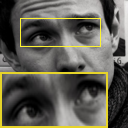}~~\\

\includegraphics[width=0.092\textwidth]{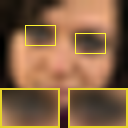}~~
		&\includegraphics[width=0.092\textwidth]{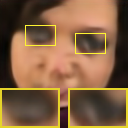}~~
		&\includegraphics[width=0.092\textwidth]{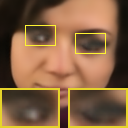}~~
		&\includegraphics[width=0.092\textwidth]{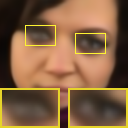}~~
        &\includegraphics[width=0.092\textwidth]{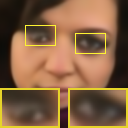}~~
		&\includegraphics[width=0.092\textwidth]{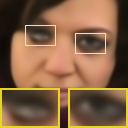}~~
		&\includegraphics[width=0.092\textwidth]{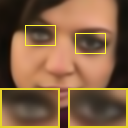}~~
        &\includegraphics[width=0.092\textwidth]{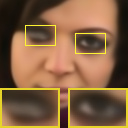}~~
        &\includegraphics[width=0.092\textwidth]{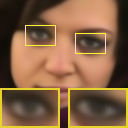}~~
        &\includegraphics[width=0.092\textwidth]{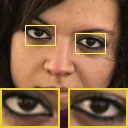}~~\\

\hspace{-0mm}
\footnotesize LR   & \footnotesize FACN~\cite{xin2020facial} & \footnotesize DIC~\cite{ma2020deep} & \footnotesize SPARNet~\cite{chen2020learning}  & \footnotesize AD-GNN~\cite{bao2022attention}  & \footnotesize Restormer-M~\cite{zamir2022restormer} & \footnotesize LAAT~\cite{LAATransformer} & \footnotesize SFMNet~\cite{wang2023spatial} & \footnotesize ~\textbf{Ours} & \footnotesize GT\\
\end{tabular}}
\vspace{-0cm}
\caption{Qualitative comparison for $\times 8$ FSR on CelebA~\cite{liu2015deep} and Helen~\cite{le2012interactive} test sets. Our method recovers detailed face images.}
\label{fig:vision}
\end{figure*}

\paragraph{Wavelet Feature Upgrade (WFU)} To obtain more details, several methods~\cite{bao2022attention,LAATransformer} propose using residual concatenation to enable the decoder to leverage information from the encoder. As the resolution of features at different scales differs, upsampling is employed to align them to the same resolution before feature fusion. Nevertheless, direct fusion operation is not optimal as it may introduce some degree of high and low-frequency aliasing. To better fuse features from the encoder, we leverage the wavelet transform for image scale transformations, developing a WFU to effectively utilize features from different scales in the decoder to enhance facial details by fusing high and low-frequency features separately.

Specifically, as shown in Fig.~\ref{fig:main}, for larger scale feature \hbox{${\boldsymbol{F}_s}$~$\in$~$\mathbb{R}^{\frac{H}{4} \times \frac{W}{4} \times 4C}$} from the encoder and smaller scale feature \hbox{${\boldsymbol{F}_{s + 1}}$~$\in$~$\mathbb{R}^{\frac{H}{8} \times \frac{W}{8} \times 4C}$} from the decoder, we initially apply the wavelet transform to the larger scale feature \hbox{${\boldsymbol{F}_s}$}, resulting in four wavelet subbands of same scale as \hbox{$\mathbb{R}^{\frac{H}{8} \times \frac{W}{8} \times 4C}$}:
\begin{equation}
\{ \boldsymbol{A}_{LL}^s,\boldsymbol{H}_{LR}^s,\boldsymbol{V}_{RL}^s,\boldsymbol{D}_{RR}^s\}  = \mathcal{WT}({\boldsymbol{F}_s}),
\end{equation}
where \hbox{$\boldsymbol{A}_{LL}^s$} represents the low-frequency part of \hbox{${\boldsymbol{F}_s}$ and $\boldsymbol{H}_{LR}^s$}, \hbox{$\boldsymbol{V}_{RL}^s$}, \hbox{$\boldsymbol{D}_{RR}^s$} represent the three high-frequency parts of \hbox{${\boldsymbol{F}_s}$}. Considering that small-scale feature \hbox{${\boldsymbol{F}_{s + 1}}$} is presumed to contain low-frequency information predominantly, we combine \hbox{$\boldsymbol{A}_{LL}^s$} with it as the enhanced low-frequency subband. Simultaneously, we employ a residual block to strengthen the high-frequency components of the image, and ultimately, output \hbox{${\boldsymbol{F}_{s}^{\prime}}$~$\in$~$\mathbb{R}^{\frac{H}{4} \times \frac{W}{4} \times 4C}$} can be obtained through the inverse wavelet transform:
\begin{equation}
\boldsymbol{F}_{s}^{\prime}= \mathcal{IWT}\left( {\mathcal{C}\left( {\boldsymbol{A}_{LL}^s,{\boldsymbol{F}_{s + 1}}} \right),\mathcal{R}\left( {\boldsymbol{H}_{LR}^s,\boldsymbol{V}_{RL}^s,\boldsymbol{D}_{RR}^s} \right)} \right),
\end{equation}
where \hbox{${\mathcal{IWT}}$} denotes the inverse wavelet transform, \hbox{$\mathcal{\mathcal{C}}$} denotes the concatenation, and \hbox{$\mathcal{R}$} denotes the standard residual block.

\subsection{Full-domain Transformer}
As analyzed above, the main path of the framework consists mainly of low-frequency information. Therefore, utilizing the Transformer structure, which exhibits greater sensitivity to low-frequency information, is more advantageous for facial feature extraction. To enhance the restoration of face images, it is crucial to effectively utilize face features at local, regional, and global levels. Specifically, local regions encompass multiple pixels and are most effectively modeled using small \hbox{$1 \times 1$} or \hbox{$3 \times 3$} kernels, capturing typical features such as local facial details. Regional features encompass dozens of pixel points, such as eyes, nose, and other facial components. Due to their larger spatial extent, they are better modeled using convolution with large kernels~\cite{ding2022scaling,sun2023spatially} or window-based Transformers~\cite{li2023efficient}. Global features involve the structural correlation of the entire face, such as the overall facial contour, and are best modeled using the global Transformer. However, most methods~\cite{zamir2022restormer,li2022efficient,wang2022faceformer,li2023cross} only focus on leveraging local and global features or local and regional features. Thus, as shown in Fig.~\ref{fig:attention}, we propose a full-domain Transformer as primary module for feature extraction, which consists of two main parts: regional self-attention (RSA) focuses on extracting local and regional facial features, while global self-attention (GSA) is responsible for extracting local and global facial features. Subsequently, we will elaborate on how FDT effectively captures local, regional, and global facial features.

\begin{table}[t!]
\tiny
\setlength\tabcolsep{4pt}
\centering
\vspace{0mm}
    \caption{Comparison of face similarity on SCface~\cite{grgic2011scface} test set.}
\vspace{-0mm}
\label{tab:scface}
\resizebox{0.48\textwidth}{!}{
\begin{tabular}{l||ccccc}
\toprule
\rowcolor{lightgray}
& \multicolumn{5}{c}{Average similarity$\uparrow$} 
\\ 
\cmidrule{2-6}
    \rowcolor{lightgray}
    \multicolumn{1}{l||}{\multirow{-2}{*}{Methods}} 
    & Case 1  & Case 2    & Case 3
    & Case 4  & Case 5    
    \\ 
    \hline\hline
    FSRNet~\cite{chen2018fsrnet}     & 0.6713   & 0.6560  & 0.6794     & 0.6903   & 0.6711    \\ 
    FACN~\cite{xin2020facial}     & 0.6545   & 0.6318  & 0.6571     & 0.6710   & 0.6516   \\ 
    DIC~\cite{ma2020deep}     & 0.5272   & 0.4851  & 0.5772     & 0.5431   & 0.5527  \\ 
    SPARNet~\cite{chen2020learning} & 0.7100   & 0.6911  & 0.7160     & 0.7252   & 0.7041   \\ 
    AD-GNN~\cite{bao2022attention}     & 0.7188   & 0.6947  & 0.7171     & 0.7283   & 0.7161   \\ 
    % Restormer-M~\cite{zamir2022restormer}    & -   & -  & -     & -   & - \\ 
    LAAT~\cite{LAATransformer}     & 0.7193   & 0.7070  & 0.7140     & \underline{0.7342}   & \underline{0.7238}   \\ 
    SFMNet~\cite{wang2023spatial}    & \underline{0.7224}   & \underline{0.7101}  & \underline{0.7243}     & 0.7331   & 0.7223 \\ 
    \bf{Ours}    &  \bf{0.7252}   & \bf{0.7239}  & \bf{0.7253}     & \bf{0.7426}   & \bf{0.7256} \\ 

    \bottomrule
\end{tabular}}
\end{table}
\hspace{-0mm}

% Ablation studies of WFD and WFU, as well as on information exchange mechanisms in the full-domain transformer, including the shift window (SW) and shuffle heads (SH) on Helen test set.

\begin{table}[t!]
\tiny
\setlength\tabcolsep{2.5pt}
\centering
\vspace{0mm}
    \caption{Ablation studies of WFD, WFU, shift window (SW) mechanism, and shuffle heads (SH) mechanism in our full-domain Transformer on Helen~\cite{le2012interactive} test set.}
\vspace{-0mm}
\label{tab:ablation}
\resizebox{0.48\textwidth}{!}{
\begin{tabular}{l||cccc|l|l|l}
\toprule
\rowcolor{lightgray}
    \rowcolor{lightgray}
    \multicolumn{1}{l||}{\multirow{-1}{*}{Methods}} 
    & \multicolumn{1}{c}{\multirow{-1}{*}{WFD}}
    & \multicolumn{1}{c}{\multirow{-1}{*}{WFU}}
    & \multicolumn{1}{c}{\multirow{-1}{*}{SW}}
    & \multicolumn{1}{c|}{\multirow{-1}{*}{SH}}
    & \multicolumn{1}{c|}{\multirow{-1}{*}{Params}}
    & \multicolumn{1}{c|}{\multirow{-1}{*}{FLOPs}}
    & \multicolumn{1}{c}{\multirow{-1}{*}{PSNR / SSIM}}

    \\ 
    \hline\hline
    % Baseline     & \XSolidBrush   & \XSolidBrush  & \XSolidBrush  & \XSolidBrush  & -  & -  & -  \\ 
    \emph{w/o} WFD     & \XSolidBrush   & \textcolor{gray}{\Checkmark}  & \textcolor{gray}{\Checkmark} & \textcolor{gray}{\Checkmark}  & 0.830M & 1.131G & 26.22 / 0.7743 \\ 
    \emph{w/o} WFU     & \textcolor{gray}{\Checkmark}   & \XSolidBrush  & \textcolor{gray}{\Checkmark} & \textcolor{gray}{\Checkmark}  & 0.719M & 1.085G & 26.27 / 0.7772 \\ 
    \emph{w/o} SW     & \textcolor{gray}{\Checkmark}   & \textcolor{gray}{\Checkmark}  & \XSolidBrush & \textcolor{gray}{\Checkmark}  & 0.848M & 1.164G & 26.31 / 0.7763 \\ 
    \emph{w/o} SH     & \textcolor{gray}{\Checkmark}   & \textcolor{gray}{\Checkmark}  & \textcolor{gray}{\Checkmark} & \XSolidBrush  & 0.848M & 1.164G & 26.31 / 0.7783 \\ 
    \textbf{Ours}     & \textcolor{gray}{\Checkmark}   & \textcolor{gray}{\Checkmark}  & \textcolor{gray}{\Checkmark} & \textcolor{gray}{\Checkmark}  & 0.848M & 1.164G & \bf{26.36 / 0.7795} \\ 

    \bottomrule
\end{tabular}}
\end{table}
\hspace{-0mm}

\paragraph{Regional Self-Attention (RSA)} For an input layer normalized facial feature \hbox{$\boldsymbol{X}$~$\in$~$\mathbb{R}^{H\times W \times C}$}, we first extract a set of window features from input \hbox{$\boldsymbol{X}$}:
\begin{equation}
\{ {\boldsymbol{X}_1},{\boldsymbol{X}_2},...,{\boldsymbol{X}_n}\}  = Split(\boldsymbol{X}),
\end{equation}
where \hbox{$\{ {\boldsymbol{X}_1},{\boldsymbol{X}_2},...,{\boldsymbol{X}_n}\} $~$\in$~$\mathbb{R}^{\frac{{HW}}{{N^2}} \times N \times N \times C}$} denotes a set of window feature patches, \hbox{$N$} denotes the size of the window, and \hbox{$n = \frac{H}{N} = \frac{W}{N}$}. Subsequently, the model initially captures the local facial details to enhance the network's contextual information. Local details are captured by combining a \hbox{$1 \times 1$} point-wise convolution and a \hbox{$3 \times 3$} depth-wise convolution. Then for each window feature patch \hbox{${\boldsymbol{X}_i}$} that enhances the local context, we project it into query \hbox{${\boldsymbol{Q}_i}$~$\in$~$\mathbb{R}^{\frac{{HW}}{{N^2}} \times {N^2} \times C}$}, key \hbox{${\boldsymbol{K}_i}$~$\in$~$\mathbb{R}^{\frac{{HW}}{{N^2}} \times {N^2} \times C}$}, and value \hbox{${\boldsymbol{V}_i}$~$\in$~$\mathbb{R}^{\frac{{HW}}{{N^2}} \times {N^2} \times C}$}. This process can be described as:
\begin{equation}
\{ {\boldsymbol{Q}_i},{\boldsymbol{K}_i},{\boldsymbol{V}_i}\}  = \mathcal{RS}\left( {\mathcal{D}\left( {\mathcal{P}\left( {{\boldsymbol{X}_i}} \right)} \right)} \right),
\end{equation}
where \hbox{$\mathcal{RS}$} denotes a reshape operator, \hbox{${\mathcal{D}}$} denotes a depth-wise convolution layer, and \hbox{${\mathcal{P}}$} denotes a point-wise convolution layer. On this basis, for each window feature patch, regional self-attention can be formulated as:
\begin{equation}
{\rm{Attention}}({\boldsymbol{Q}_i},{\boldsymbol{K}_i},{\boldsymbol{V}_i}) = {\boldsymbol{V}_i}{\rm{ReLU}}({\boldsymbol{Q}_i}{\boldsymbol{K}_i}^T/\alpha ),
\end{equation}
Here, \hbox{$\alpha $} denotes a learnable parameter. To avoid the computational complexity of \hbox{$\mathcal{O}(H^2W^2)$}, we choose to implicitly encode global features across channels when computing the feature covariance \hbox{$\boldsymbol{A}$}. Specifically, we replace the attention map of size \hbox{$\boldsymbol{A}$~$\in$~$\mathbb{R}^{HW\times HW}$} in the traditional sense with a regional attention map of size \hbox{$\boldsymbol{A}$~$\in$~$\mathbb{R}^{C\times C}$}. Furthermore, to address the absence of connectivity among different windows, as shown in Fig.~\ref{fig:attention}, we introduce a straightforward yet effective information exchange mechanism for RSA to facilitate communication between adjacent windows by shifting windows. Hence, our meticulous design allows RSA to enhance regional and local facial features effectively.

\paragraph{Global Self-Attention (GSA)} For an input layer normalized facial feature \hbox{${\boldsymbol{X}^{\prime}}$~$\in$~$\mathbb{R}^{H\times W \times C}$}, similarly, we initially employ \hbox{$1 \times 1$} point-wise convolution and \hbox{$3 \times 3$} depth-wise convolution to extract local information from \hbox{${\boldsymbol{X}^{\prime}}$}, ensuring the accurate recovery of facial details. Subsequently, we adhere to prior methods~\cite{zamir2022restormer} by subdividing the channel into multi-heads \hbox{$h$} and concurrently learning distinct self-attention maps. Specifically, we generate query \hbox{$\boldsymbol{Q}$~$\in$~$\mathbb{R}^{h \times {\hat C} \times HW}$}, key \hbox{$\boldsymbol{K}$~$\in$~$\mathbb{R}^{h \times {\hat C} \times HW}$}, value \hbox{$\boldsymbol{V}$~$\in$~$\mathbb{R}^{h \times {\hat C} \times HW}$} projections based on the overall face feature after enhanced local detail, where \hbox{$\hat C$} is the number of channels in each head. Next, we create a global attention map of size \hbox{$\boldsymbol{A}^{\prime}$~$\in$~$\mathbb{R}^{{\hat C}   \times {\hat C}}$} by computing the dot product of vectors \hbox{$\boldsymbol{Q}$} and \hbox{$\boldsymbol{K}$}. This process emphasizes the relationships between channels while implicitly encoding global facial features. In summary, the full process of global self-attention can be formulated as follows: 
\begin{equation}
{\rm{Attention}}(\boldsymbol{Q},\boldsymbol{K},\boldsymbol{V}) = \boldsymbol{V}{\rm{ReLU}}(\boldsymbol{Q}{\boldsymbol{K}^T}/\beta  ),
\end{equation}
where \hbox{$\beta $} denotes a learnable parameter. To augment information exchange between the multi-heads, as illustrated in Fig.~\ref{fig:attention}, we achieve this by blending multi-head feature mechanisms. Through meticulous design, our GSA effectively enhances the local and global features of the input face images.

\begin{figure}
\begin{overpic}[width=0.99\linewidth]{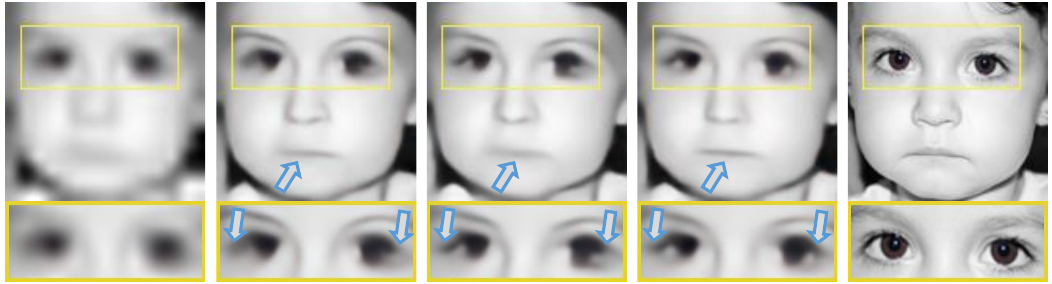}
\put(6.5,-3.8){\color{black}{\fontsize{7pt}{1pt}\selectfont (a) LR}}
\put(22.5,-3.8){\color{black}{\fontsize{7pt}{1pt}\selectfont (b) \emph{w/o} WFD}}
 \put(42.5,-3.8){\color{black}{\fontsize{7pt}{1pt}\selectfont (c) \emph{w/o} WFU}}
 \put(64.5,-3.8){\color{black}{\fontsize{7pt}{1pt}\selectfont (d) \bf{Ours}}}
  \put(86.0,-3.8){\color{black}{\fontsize{7pt}{1pt}\selectfont (e) GT}}
\end{overpic}
\vspace{3.5mm}
   \caption{Effectiveness of our WFD and WFU. We use bicubic downsample and upsample with comparable parameters instead of WFD and WFU, respectively.}
\label{fig:up_down}
\vspace{-0mm}
\end{figure}

\begin{figure}
\begin{overpic}[width=0.99\linewidth]{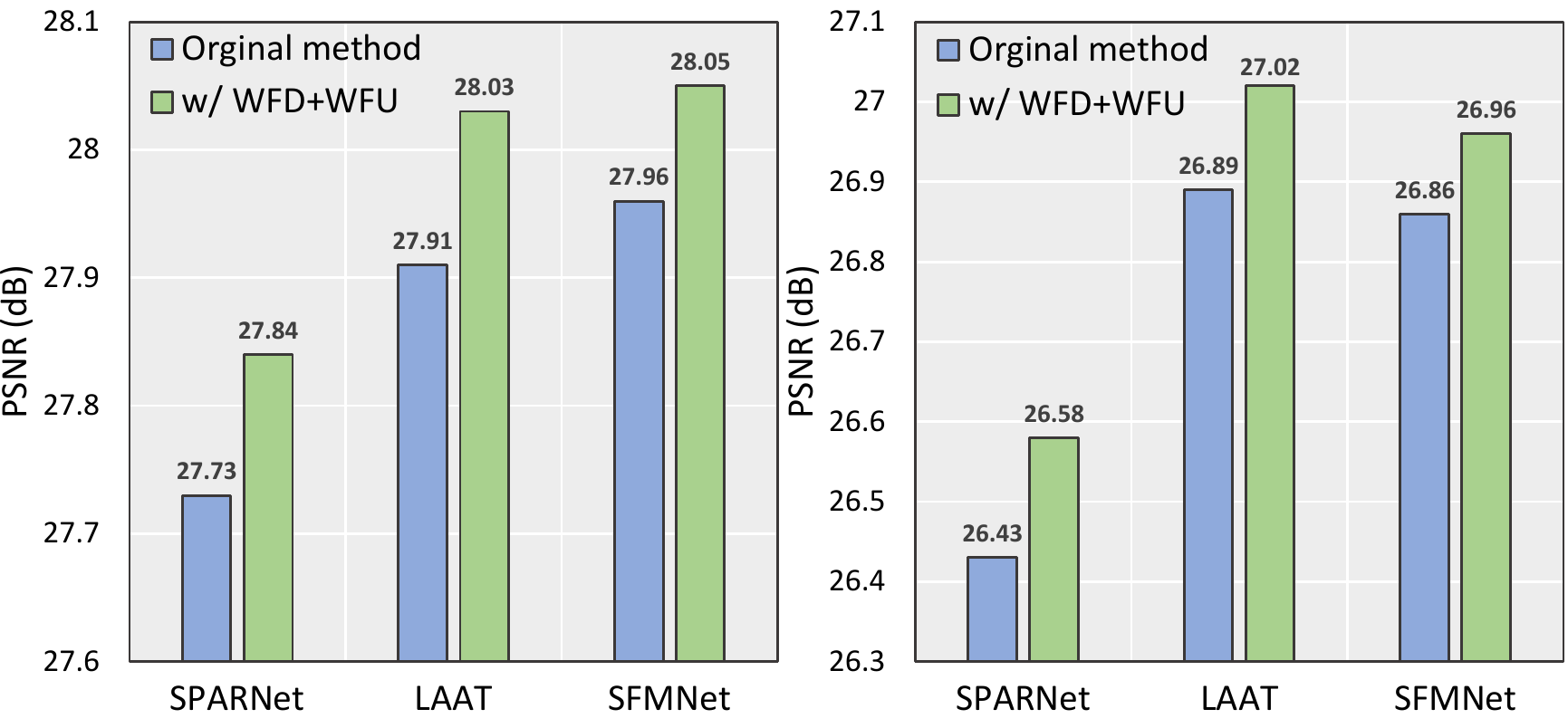}
\put(12,-4.4){\color{black}{\fontsize{7.5pt}{1pt}\selectfont (a) On the CelebA test set.}}
\put(63,-4.4){\color{black}{\fontsize{7.5pt}{1pt}\selectfont (b) On the Helen test set.}}
\end{overpic}
\vspace{5mm}
   \caption{Ablation studies on the generalizability of WFD+WFU. We add WFD+WFU to SPARNet~\cite{chen2020learning}, LAAT~\cite{LAATransformer}, and SFMNet~\cite{wang2023spatial} and observe PSNR enhancement.}
\label{fig:ablation}
\vspace{-0mm}
\end{figure}

\begin{figure*}[ht]
\begin{overpic}[width=0.99\linewidth]{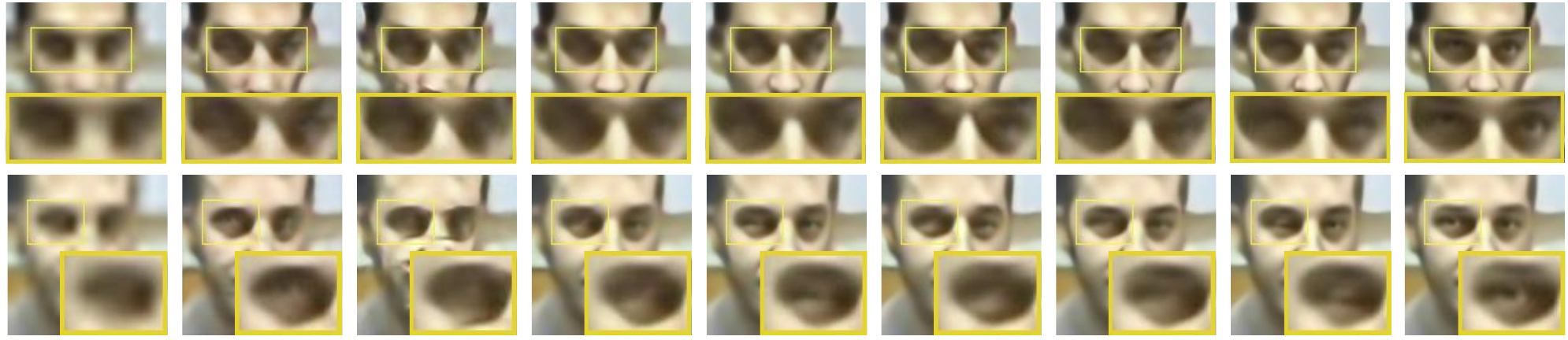}
\put(4.5,-1.8){\color{black}{\fontsize{8pt}{1pt}\selectfont LR}}
\put(13.1,-1.8){\color{black}{\fontsize{8pt}{1pt}\selectfont FSRNet~\cite{chen2018fsrnet}}}
 \put(24.5,-1.8){\color{black}{\fontsize{8pt}{1pt}\selectfont FACN~\cite{xin2020facial}}}
 \put(34.9,-1.8){\color{black}{\fontsize{8pt}{1pt}\selectfont SPARNet~\cite{chen2020learning}}}
  \put(45.8,-1.8){\color{black}{\fontsize{8pt}{1pt}\selectfont AD-GNN~\cite{bao2022attention}}}
  \put(55.5,-1.8){\color{black}{\fontsize{8pt}{1pt}\selectfont Restormer-M~\cite{zamir2022restormer}}}
  \put(69.0,-1.8){\color{black}{\fontsize{8pt}{1pt}\selectfont LAAT~\cite{LAATransformer}}}
  \put(79.6,-1.8){\color{black}{\fontsize{8pt}{1pt}\selectfont SFMNet~\cite{wang2023spatial}}}
  \put(93.0,-1.8){\color{black}{\fontsize{8pt}{1pt}\selectfont \textbf{Ours}}}
\end{overpic}
\vspace{4mm}
   \caption{Qualitative comparison of the state-of-the-art methods on SCface~\cite{grgic2011scface} test set. Our method can restore clear face components, especially the eye region, which is critical for downstream face recognition tasks.}
\label{fig:scface_vision}
\vspace{-0.2mm}
\end{figure*}

\section{Experiments}
\subsection{Datasets and Evaluation Metrics}
We employ the CelebA~\cite{liu2015deep} dataset for training and evaluate the models on the CelebA~\cite{liu2015deep}, Helen~\cite{le2012interactive}, and SCface~\cite{grgic2011scface} datasets. Due to variations in the length and width of the original face image, we pre-detect the 68 landmarks of the face using OpenFace~\cite{baltrusaitis2018openface}. The face images are then cropped based on these landmarks and resized to 128$ \times $128 pixels to serve as the ground truth. The ground truth images are further downsampled to 16$ \times $16 to generate LR images using bicubic interpolation. Based on this foundation, we utilize 18,000 samples from CelebA for training. For testing, we select 1,000 samples from CelebA and 50 samples from Helen. As for the quality assessment metrics, we used PSNR, SSIM~\cite{wang2002universal}, LPIPS~\cite{zhang2018unreasonable} and VIF~\cite{sheikh2006image}. Recognizing the significance of identity consistency, we introduced the identity comparison accuracy, denoted as ID. This metric uses SFace~\cite{zhong2021sface} as a recognition model, determining whether the restored and original faces belong to the same identity.

\begin{table}[t!]
\tiny
\setlength\tabcolsep{1.5pt}
\centering
\vspace{0mm}
    \caption{Ablation studies on the efficiency of our full-domain Transformer (FDT). We use FDT to replace feature extraction modules in SPARNet~\cite{chen2020learning}, Restormer~\cite{zamir2022restormer}, and LAAT~\cite{LAATransformer}, respectively. Our FDT can achieve gained performance with reduced computational costs.}
\vspace{-0mm}
\label{tab:FDT}
\resizebox{0.48\textwidth}{!}{
\begin{tabular}{l||l|l|c}
\toprule
\rowcolor{lightgray}
& & & \multicolumn{1}{c}{Helen~\cite{le2012interactive}} 
\\ 
\cmidrule{4-4}
    \rowcolor{lightgray}
    \multicolumn{1}{l||}{\multirow{-2}{*}{Methods}} 
    % & \multicolumn{1}{c|}{\multirow{-2}{*}{Speed}}
    & \multicolumn{1}{c|}{\multirow{-2}{*}{Params}}
    & \multicolumn{1}{c|}{\multirow{-2}{*}{FLOPs}}
    & PSNR / SSIM / LPIPS$\downarrow$ / VIF$\uparrow$

    \\ 
    \hline\hline

    Ours+SPARNet~\cite{chen2020learning}       & 0.925M & 2.565G  & 26.27 / 0.7754 / 0.2804 / 0.4241\\ 

    Ours+Restormer~\cite{zamir2022restormer}      & 1.063M & 1.663G  & 26.33 / 0.7770 / \underline{0.2747} / \underline{0.4259}\\ 

    Ours+LAAT~\cite{LAATransformer}      & 1.089M & 1.863G  & \underline{26.33} / \underline{0.7771} / 0.2801 / 0.4250\\ 

    \textbf{Ours+FDT}      & \bf{0.848M} & \bf{1.164G} & \bf{26.36} / \bf{0.7795} / \bf{0.2745} / \bf{0.4283}\\ 

    \bottomrule
\end{tabular}}
\end{table}
\hspace{-0mm}

% We implement all experiments using the PyTorch framework with a single NVIDIA GeForce RTX 4090. 

\subsection{Implementation Details}
In the network, we first extend the number of channels \hbox{$C$} to 40. And in \hbox{$\mathbb{R}^{H\times W \times C}$} stage, the number of full-domain Transformer is set to 2, in \hbox{$\mathbb{R}^{\frac{H}{8} \times \frac{W}{8} \times 4C}$} stage the number is set to 6, and in all the remaining stages the number is set to 1. During the training stage, our model is optimized with an L1 loss with a coefficient of 1, and we use the Adam optimizer with \hbox{${\beta _{\rm{1}}}$}=0.9, \hbox{${\beta _{\rm{2}}}$}=0.99. We set the initial learning rate to \hbox{${\rm{2}} \times {e^{ - 4}}$}. The ID metric's cosine threshold for identity matching is set to 0.5 in TABLE~\ref{tab:performance}.

\subsection{Comparisons with State-of-the-Art Methods}
We benchmark our method against several state-of-the-art FSR methods using a unified dataset. The compared methods include prior-based approaches like FSRNet~\cite{chen2018fsrnet}, FACN~\cite{xin2020facial}, and DIC~\cite{ma2020deep}, attention-based CNN methods such as SPARNet~\cite{chen2020learning} and AD-GNN~\cite{bao2022attention}, and Transformer-based methods like Restormer-M~\cite{zamir2022restormer}, LAAT~\cite{LAATransformer}, and SFMNet~\cite{wang2023spatial}, where Restormer-M is a generalized image restoration method fine-tuned on face training sets. We present the quantitative results for the CelebA and Helen test datasets in TABLE~\ref{tab:performance}. The best and the second-best results are emphasized in bold and underlined in this paper. Our method excels in various metrics, including image structure similarity (PSNR and SSIM), visual quality (LPIPS), fidelity (VIF), and face identity consistency (ID), achieving the best performance. Furthermore, we provide quantitative data about models, including the number of model parameters, FLOPs, and inference speed, in TABLE~\ref{tab:performance} to assess the model's efficiency. Compared with the methods above, our method is less parametric and computationally intensive, and faster in inference, exhibiting excellent efficiency. Next, we visually compare the restoration results of various methods. As shown in Fig.~\ref{fig:vision}, the high-frequency face profile achieved by our method is significantly sharper and more closely resembles the ground truth, such as key facial components such as the eyes. More qualitative comparisons can also be found in \emph{supplementary materials}.

% \put(4.1,-2){\color{black}{\fontsize{8pt}{1pt}\selectfont (a) PSNR on Helen test set during training.}}
% \put(37.5,-2){\color{black}{\fontsize{8pt}{1pt}\selectfont (b) PSNR on CelebA test set during training.}}
% \put(72,-2){\color{black}{\fontsize{8pt}{1pt}\selectfont (c) Comparison of computational loads.}}

\begin{figure*}
\begin{overpic}[width=0.99\linewidth]{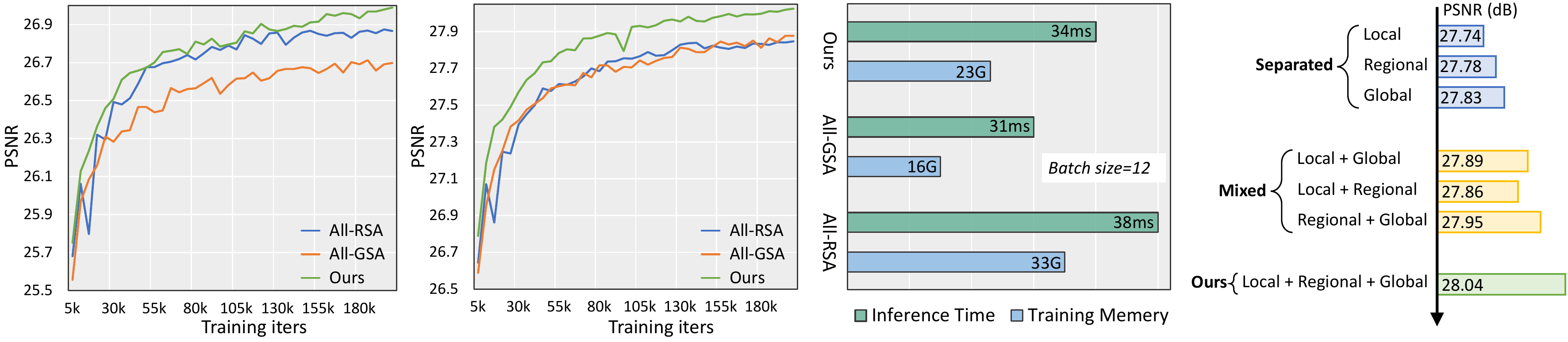}
\put(2.1,-2){\color{black}{\fontsize{7pt}{1pt}\selectfont (a) Training PSNR on Helen~\cite{le2012interactive} test set.}}
\put(27.2,-2){\color{black}{\fontsize{7pt}{1pt}\selectfont (b) Training PSNR on CelebA~\cite{liu2015deep} test set.}}
\put(52.5,-2){\color{black}{\fontsize{7pt}{1pt}\selectfont (c) Comparison of computational loads.}}
\put(81.5,-2){\color{black}{\fontsize{7pt}{1pt}\selectfont (d)  Comparison of PSNR.}}
\end{overpic}
\vspace{3mm}
   \caption{Ablation studies on the effectiveness of local, regional, and global facial features for FSR.}
\label{fig:Transformer}
\vspace{-0mm}
\end{figure*}

Additionally, we validate the efficacy of our method in a practical surveillance scenario. For this purpose, we chose HR face images of test subjects from the SCface dataset as the source samples. The corresponding LR face images captured by surveillance cameras are regarded as the target samples. We created five case groups, each consisting of 5 pairs of randomly selected face samples. The evaluation metric is the average similarity between the restored and HR faces. As shown in TABLE~\ref{tab:scface}, our method consistently reconstructs faces with higher similarity in all cases, which indicates that our method can be better applied to a practical scenario. In addition, visual comparisons on the SCface test set of various methods can be found in Fig.~\ref{fig:scface_vision}, where prior-based FSR methods exhibit varying degrees of distortion in key facial components. This distortion could be attributed to inaccurate prior estimation, particularly at the current very low resolutions. Attention-based and Transformer-based methods improved the clarity of the restored face to some extent, but the face contours and edges were still not clear. In contrast, our method excels at restoring the contours of the face and facial components with superior clarity, a crucial aspect for downstream tasks like face matching. In summary, the comprehensive results, both quantitative and qualitative, illustrate the efficiency of our model's performance as well as its applicability across various scenarios. 

\subsection{Ablation Study}
% This subsection presents an experimental ablation analysis of the causes of our method's effectiveness, including two reasons: the proposed wavelet-based downsample and upgrade modules and the proposed full-domain Transformer.

\paragraph{Wavelet Feature Downsample and Upgrade.} WFD and WFU are important components in our wavelet-based encoder-decoder structure. To assess the efficacy of our proposed WFD and WFU, we conducted experiments by substituting WFD with stride convolution for downsampling and WFU with interpolation for upsampling. As indicated in TABLE~\ref{tab:ablation}, the computational burden imposed by the WFD module for downsampling is nearly negligible. However, leveraging the WFD for downsampling significantly enhances the model's performance, resulting in a noteworthy PSNR gain of 0.14 dB. Subsequently, we observe that employing WFU to enhance facial details in the decoder stage yields a modest performance improvement compared to the conventional interpolation method of upsampling. This enhancement leads to a PSNR gain of 0.09 dB while maintaining a relatively modest computational load. Meanwhile, corresponding visual comparisons are presented in Fig.~\ref{fig:up_down}. In Fig.~\ref{fig:up_down} (b), without using WFD, the contours around the eyes, mouth, and corners of the face appear somewhat blurred. Similarly, in Fig.~\ref{fig:up_down} (c), without using WFU, the contours around the right eye corner and mouth appear blurred. In contrast, Fig.~\ref{fig:up_down} (d) with the complete WFD and WFU, reconstructed face component contours are sharper and closer to ground truth. This portion of the experiment strongly demonstrates the effectiveness of our WFD and WFU.

Subsequently, to assess the generalization of WFD plus WFU, we incorporate both into several existing methods, replacing their native downsampling and upsampling while preserving their proposed feature extraction modules. These methods include SPARNet~\cite{chen2020learning}, LAAT~\cite{LAATransformer} and SFMNet~\cite{wang2023spatial}. As depicted in Fig.~\ref{fig:ablation}, all these methods exhibit significant performance enhancements when integrated with WFD plus WFU, with PSNR gains greater than 0.1dB. Notably, our proposed WFD and WFU are remarkably lightweight, imposing an almost negligible additional computational load. In summary of the two-part ablation studies presented above, WFD and WFU are efficient downsampling and upsampling approaches that can be seamlessly integrated into existing methods.

\paragraph{Full-domain Transformer.} 
To assess the impact of extracting local, regional, and global facial features on facial reconstruction, we replace the combinations of RSA and GSA in FDT with all-RSA or all-GSA, respectively. This simulates existing FSR methods that exclusively focus on global and local or only regional and local facial features. Meanwhile, it can ensure that the calculated loads of the three ablation methods are comparable for a fair comparison. As depicted in Fig.~\ref{fig:Transformer} (a) and (b), our full-domain Transformer exhibits faster training convergence and superior performance on both test sets. Additionally, as illustrated in Fig.~\ref{fig:Transformer} (c), our full-domain Transformer exhibits a balanced computational load, including inference speed and training memory. To further showcase the effectiveness of simultaneously capturing local, regional, and global features of a face image, we illustrate the separated case, mixed case, and our full-domain case in Fig.~\ref{fig:Transformer} (d). The separated case refers to situations where only one of the local, regional, or global features of the face image is focused on. Mixed case refers to situations where two of the three facial features-local, regional, and global-are attended to. In contrast to the above two cases, our method achieves a notable performance enhancement by incorporating complementary features across various scales, making it the optimal solution. All experimental results show that simultaneously focusing on local, regional, and global features of face images can effectively enhance performance without significantly increasing costs. 

Next, TABLE~\ref{tab:ablation} substantiates the significance of information exchange mechanisms, including exchanging facial information across distinct regions via shifting windows in RSA and exchanging different multi-head information via shuffling heads in GSA. As indicated in TABLE~\ref{tab:ablation}, these information exchange mechanisms incur almost no computational cost, yet incorporating both mechanisms separately results in a PSNR gain of 0.05 dB in the model's performance. Moreover, we determine the efficiency of FDT by using the basic feature extraction modules in the three methods, SPARNet~\cite{chen2020learning}, Restormer~\cite{zamir2022restormer}, and LAAT~\cite{LAATransformer}, instead of FDT. As can be seen from TABLE~\ref{tab:FDT}, FDT achieves the best performance in several metrics with fewer numbers of parameters and FLOPs compared to these modules. This result fully demonstrates that the proposed FDT is a more efficient module to deal with FSR. Therefore, with our proposed FDT as the main feature extraction module, our method has a more powerful feature extraction capability than existing methods. More ablation studies can also be found in \emph{supplementary materials}.

% \paragraph{Discussion.} Experimental results show that the efficiency of our method is contributed by two parts: wavelet-based encoder-decoder structure and full-domain Transformer. Both parts can be integrated into existing encoder-decoder-based methods~\cite{chen2020learning,wang2023spatial,LAATransformer} to further enhance their performance. Therefore, our method is not only efficient in performance but also generalizable.

\section{Conclusion}
This paper presents a wavelet-based feature enhancement network for efficient FSR. To address the feature distortion caused by direct downsampling in the encoder-decoder structure, we integrate WFD and WFU into the encoder-decoder structure. Additionally, by further employing our FDT to extract low-frequency facial features comprehensively, our method can achieve a more accurate reconstruction of facial structures. We verify the effectiveness of WFD and WFU in minimizing facial structure distortion during reconstruction and the comprehensive facial feature perception capability provided by FDT. Extensive experiments, including face matching in surveillance scenarios, demonstrate that our method effectively achieves FSR with higher fidelity, achieving an excellent balance between performance, model size, and inference speed.

\section{Acknowledgments}
This work was supported in part by Beijing Natural Science Foundation Project No. Z200002, National Natural Science Foundation of China (NSFC) No. 62106022, 62225601, U23B2052, Youth Innovative Research Team of BUPT No. 2023YQTD02.

\bibliographystyle{ACM-Reference-Format}
\balance
\bibliography{camera-ready.bbl}

\end{sloppypar}
\end{document}

% \raggedbottom
\begin{sloppypar}

\title{Supplementary Material for "Efficient Face Super-Resolution via Wavelet-based Feature Enhancement Network"}
\renewcommand{\shorttitle}{Efficient Face Super-Resolution via Wavelet-based Feature Enhancement Network}

%     <ccs2012>
%     <concept>
%     <concept_id>10010147.10010178.10010224.10010245.10010254</concept_id>
%     <concept_desc>Computing methodologies~Reconstruction</concept_desc>
%     <concept_significance>500</concept_significance>
%     </concept>
%     <concept>
%     <concept_id>10010147.10010371.10010382.10010383</concept_id>
%     <concept_desc>Computing methodologies~Image processing</concept_desc>
%     <concept_significance>300</concept_significance>
%     </concept>
%     <concept>
%     <concept_id>10010147.10010178.10010224.10010226.10010236</concept_id>
%     <concept_desc>Computing methodologies~Computational photography</concept_desc>
%     <concept_significance>100</concept_significance>
%     </concept>
%     </ccs2012>

% \ccsdesc[500]{Computing methodologies~Reconstruction}
% \ccsdesc[300]{Computing methodologies~Image processing}
% \ccsdesc[100]{Computing methodologies~Computational photography}

% \keywords{Image super-resolution, Efficient transformer, Lightweight network, Model deployment}

\maketitle

\begin{figure}
\begin{center}
\includegraphics[width=0.99\linewidth]{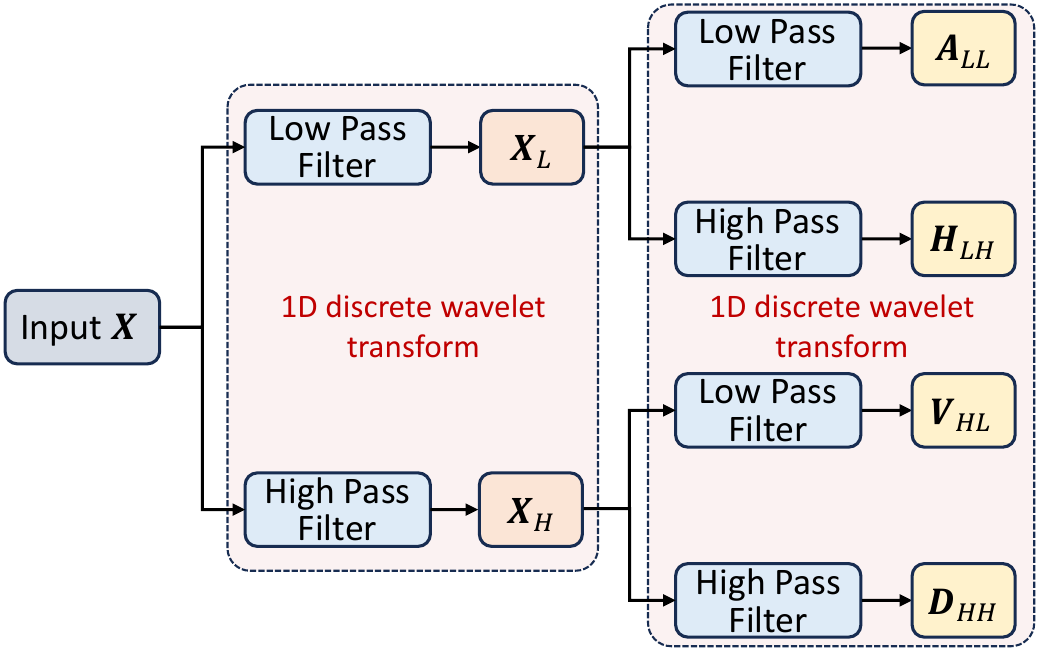}
\vspace{-0mm}
\end{center}
   \caption{Processing of 2D discrete wavelet transform (2D-DWT), where consists of two 1D discrete wavelet transform (1D-DWT).}
\label{fig:2Dwavelet}
\vspace{-0mm}
\end{figure}

\begin{figure}
\begin{center}
\includegraphics[width=0.99\linewidth]{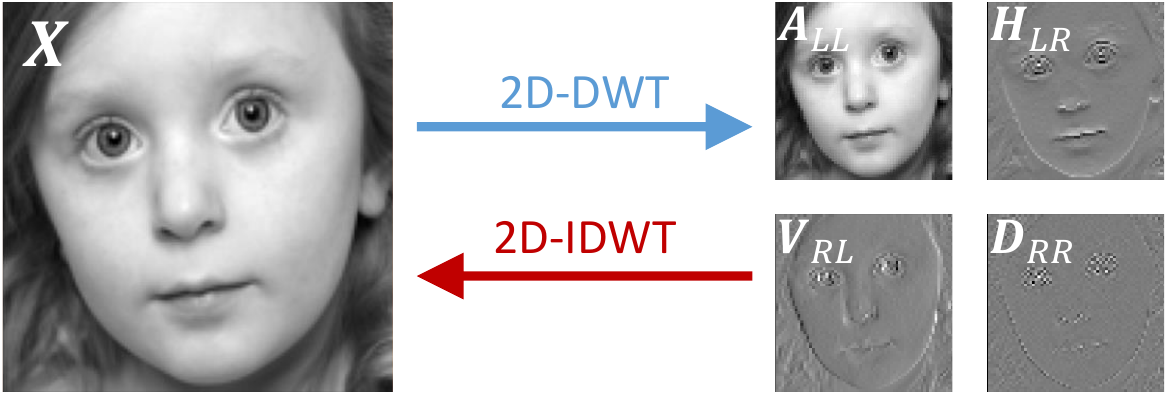}
\vspace{-0mm}
\end{center}
   \caption{2D discrete wavelet transform (2D-DWT) and 2D inverse discrete wavelet transform (2D-IDWT) processes are applied to a face image. The 2D-DWT decomposes the input face image $\boldsymbol{X}$ into one low-frequency component $\boldsymbol{A}_{LL}$ and three high-frequency components $\{ \boldsymbol{H}_{LR},\boldsymbol{V}_{RL},\boldsymbol{D}_{RR}\} $. The 2D-IDWT reduces these frequency domain components back to the original face image $\boldsymbol{X}$.}
\label{fig:2Dwavelet_show}
\vspace{-0mm}
\end{figure}

\section{2D Discrete Wavelet Transform}
2D discrete wavelet transform~\cite{mallat1989theory} (2D-DWT) is a mathematical signal processing technique employed to decompose an image into wavelet components of varying frequencies. This technique allows the analysis of image features at different frequency scales. The principle of 2D-DWT is illustrated in Fig.~\ref{fig:2Dwavelet}, and we explain the process using the commonly used Haar wavelet transform. 2D-DWT can be conceptualized as applying a 1D wavelet transform (1D-DWT) to the image using a filter in both the row and column directions. Assuming the size of the input face image $\boldsymbol{X}$~$\in$~$\mathbb{R}^{H\times W}$, where $\boldsymbol{X}\left( {i,j} \right)$ represents the pixel at position $\left( {i,j} \right)$. Initially, 1D-DWT is applied to $\boldsymbol{X}$ in the row direction, resulting in:
\begin{equation}
\boldsymbol{X}_L\left( {i,j} \right) = \mathcal{L}(k)\sum\limits_{k = 1}^{2j} \boldsymbol{X} \left( {i,2j - k} \right),
\end{equation}
\begin{equation}
\boldsymbol{X}_H\left( {i,j} \right) = \mathcal{H}(k)\sum\limits_{k = 1}^{2j} \boldsymbol{X} \left( {i,2j - k} \right),
\end{equation}
Where $\mathcal{L}$ represents low-pass filter, $\mathcal{L} = [1, 1]$, $\mathcal{H}$ represents high-pass filter, $\mathcal{H} = [1, - 1]$, $\boldsymbol{X}_L$~$\in$~$\mathbb{R}^{\frac{H}{2} \times W}$ is row low-frequency component after 1D-DWT, and $\boldsymbol{X}_H$~$\in$~$\mathbb{R}^{\frac{H}{2} \times W}$ is row high-frequency component after 1D-DWT. Next, 1D-DWT on columns is performed on $\boldsymbol{X}_L$ and $\boldsymbol{X}_H$ to get output:
\begin{equation}
\boldsymbol{A}_{LL}\left( {i,j} \right) = \mathcal{L}(k)\sum\limits_{k = 1}^{2i} \boldsymbol{X}_L \left( {2i - k,j} \right),
\end{equation}
\begin{equation}
\boldsymbol{H}_{LH}\left( {i,j} \right) = \mathcal{H}(k)\sum\limits_{k = 1}^{2i} \boldsymbol{X}_L \left( {2i - k,j} \right),
\end{equation}
\begin{equation}
\boldsymbol{V}_{HL}\left( {i,j} \right) = \mathcal{L}(k)\sum\limits_{k = 1}^{2i} \boldsymbol{X}_H \left( {2i - k,j} \right),
\end{equation}
\begin{equation}
\boldsymbol{D}_{HH}\left( {i,j} \right) = \mathcal{H}(k)\sum\limits_{k = 1}^{2i} \boldsymbol{X}_H \left( {2i - k,j} \right),
\end{equation}
where $\boldsymbol{A}_{LL}$~$\in$~$\mathbb{R}^{\frac{H}{2} \times \frac{W}{2}}$ denotes the low-frequency part of the image, which contains the overall structure and general shape information in the image, $\boldsymbol{H}_{LH}$~$\in$~$\mathbb{R}^{\frac{H}{2} \times \frac{W}{2}}$ denotes the high-frequency information in the horizontal direction of the image, including the horizontal variation of edges and details, $\boldsymbol{V}_{HL}$~$\in$~$\mathbb{R}^{\frac{H}{2} \times \frac{W}{2}}$ denotes the high-frequency information in the vertical direction of the image, including the vertical variation of edges and details, and $\boldsymbol{D}_{HH}$~$\in$~$\mathbb{R}^{\frac{H}{2} \times \frac{W}{2}}$ denotes the high-frequency information in the diagonal direction of the image, including the diagonal variation of edges and details. 

On the contrary, 2D inverse discrete wavelet transform (2D-IDWT) diminishes the four frequency-domain components $\{ \boldsymbol{X}_{LL},\boldsymbol{X}_{LH},\boldsymbol{X}_{HL},\boldsymbol{X}_{HH}\} $ back to the input $\boldsymbol{X}$, making the entire process of 2D-DWT and 2D-IDWT closed and lossless. Fig.~\ref{fig:2Dwavelet_show} shows an example of 2D-DWT and 2D-IDWT performed on a face image, where the meaning of the variables is consistent with the above formulation.

\begin{figure*}[ht]
\begin{overpic}[width=0.96\linewidth]{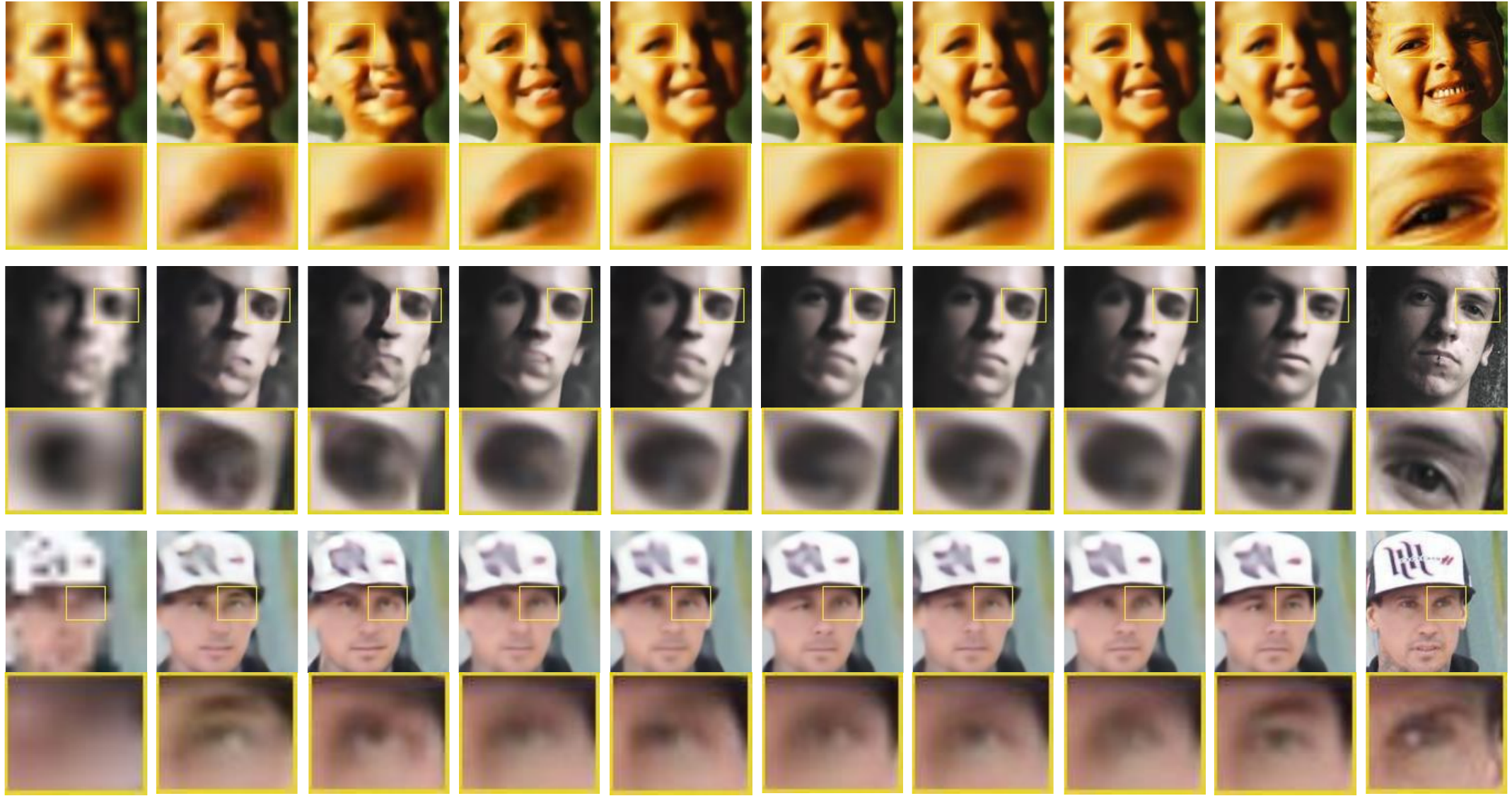}
\put(4.5,-1.8){\color{black}{\fontsize{8pt}{1pt}\selectfont LR}}
\put(11.1,-1.8){\color{black}{\fontsize{8pt}{1pt}\selectfont FSRNet~\cite{chen2018fsrnet}}}
 \put(21.5,-1.8){\color{black}{\fontsize{8pt}{1pt}\selectfont FACN~\cite{xin2020facial}}}
 \put(32.9,-1.8){\color{black}{\fontsize{8pt}{1pt}\selectfont DIC~\cite{ma2020deep} }}
  \put(40.8,-1.8){\color{black}{\fontsize{8pt}{1pt}\selectfont SPARNet~\cite{chen2020learning}}}
  \put(50.2,-1.8){\color{black}{\fontsize{8pt}{1pt}\selectfont Restormer~\cite{zamir2022restormer}}}
  \put(62.0,-1.8){\color{black}{\fontsize{8pt}{1pt}\selectfont LAAT~\cite{LAATransformer}}}
  \put(71.0,-1.8){\color{black}{\fontsize{8pt}{1pt}\selectfont SFMNet~\cite{wang2023spatial}}}
  \put(83.2,-1.8){\color{black}{\fontsize{8pt}{1pt}\selectfont \textbf{Ours}}}
  \put(93.4,-1.8){\color{black}{\fontsize{8pt}{1pt}\selectfont GT}}
\end{overpic}
\vspace{4mm}
   \caption{Qualitative comparison for $\times 8$ FSR on CelebA~\cite{liu2015deep} and Helen~\cite{le2012interactive} test datasets. Our method recovers face images that are closer to ground truth and contain more facial details than existing methods.}
\label{fig:ch_vision}
\end{figure*}

\begin{figure*}[ht]
\begin{overpic}[width=0.96\linewidth]{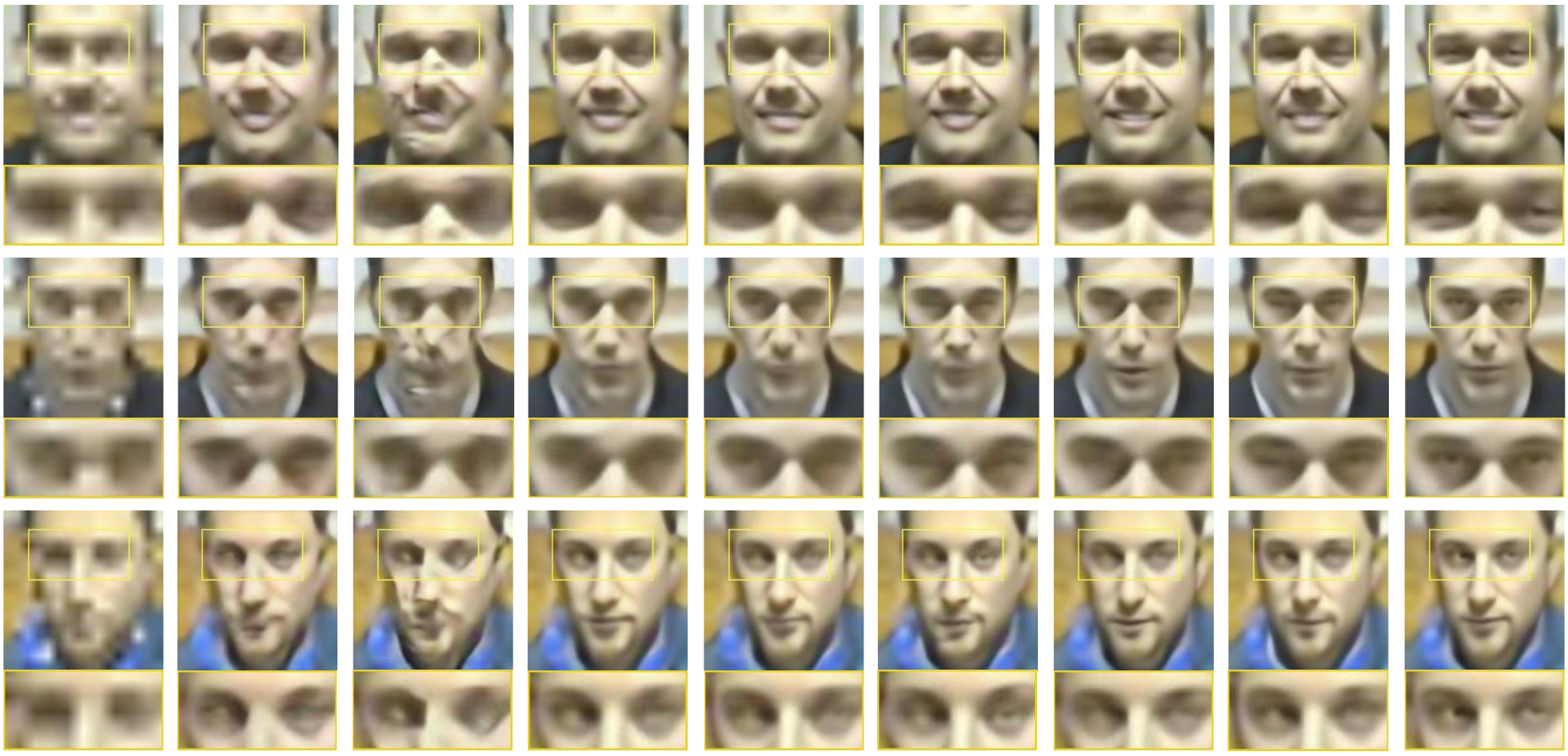}
\put(4.5,-1.8){\color{black}{\fontsize{8pt}{1pt}\selectfont LR}}
\put(13.1,-1.8){\color{black}{\fontsize{8pt}{1pt}\selectfont FSRNet~\cite{chen2018fsrnet}}}
 \put(24.5,-1.8){\color{black}{\fontsize{8pt}{1pt}\selectfont FACN~\cite{xin2020facial}}}
 \put(34.5,-1.8){\color{black}{\fontsize{8pt}{1pt}\selectfont SPARNet~\cite{chen2020learning} }}
  \put(46.4,-1.8){\color{black}{\fontsize{8pt}{1pt}\selectfont ADGNN~\cite{bao2022attention}}}
  \put(56.2,-1.8){\color{black}{\fontsize{8pt}{1pt}\selectfont Restormer~\cite{zamir2022restormer}}}
  \put(69.0,-1.8){\color{black}{\fontsize{8pt}{1pt}\selectfont LAAT~\cite{LAATransformer}}}
  \put(79.6,-1.8){\color{black}{\fontsize{8pt}{1pt}\selectfont SFMNet~\cite{wang2023spatial}}}
  \put(93.4,-1.8){\color{black}{\fontsize{8pt}{1pt}\selectfont \textbf{Ours}}}
\end{overpic}
\vspace{4mm}
   \caption{Qualitative comparison for $\times 8$ FSR on SCface~\cite{grgic2011scface} test dataset. Our method can restore the clearer face components, especially the eye region, which is critical for downstream face recognition tasks.}
\label{fig:ch_scface}
\end{figure*}

\section{More Qualitative Comparisons}
In this section, we additionally add a series of qualitative comparisons with existing methods, including prior-based methods like FSRNet~\cite{chen2018fsrnet}, FACN~\cite{xin2020facial} and DIC~\cite{ma2020deep}, attention-based methods like SPARNet~\cite{chen2020learning} and AD-GNN~\cite{bao2022attention}, as well as Transformer-based methods such as Restormer~\cite{zamir2022restormer}, LAAT~\cite{LAATransformer} and SFMNet~\cite{wang2023spatial}. Specifically, in Fig~\ref{fig:ch_vision}, we give more qualitative comparisons for $\times 8$ FSR on CelebA~\cite{liu2015deep} and Helen~\cite{le2012interactive} test datasets. In the facial region where the eyes are located, our method excels in recovering finer details such as the pupil of the eye. Moreover, the details we restore are closer to ground truth than existing methods.

In Fig~\ref{fig:ch_scface}, we give more qualitative comparisons for $\times 8$ FSR on SCface~\cite{grgic2011scface} test datasets. In the current scenario with high levels of blurring, prior-based methods exhibit poor performance, showing severe distortion. Transformer-based methods fare better by recovering general contours, yet they struggle with finer details like eye pupils. Our method excels in recovering above facial details, facilitating enhanced accuracy in downstream tasks such as recognition, significantly outperforming existing methods.

\begin{table}[t!]
\tiny
\setlength\tabcolsep{1pt}
\centering
\vspace{0mm}
    \caption{Ablation studies of down-sampling mechanisms in our method on Helen~\cite{le2012interactive} test dataset, where ``Stride'' denotes stride convolution, ``Avgpool'' denotes average pool, ``Bicubic'' denotes bicubic downsample, and ``WFD'' denotes our proposed wavelet feature downsample module. All downsampling scales are set to 2.}
\vspace{-0mm}
\label{tab:supp_ablation}
\resizebox{0.48\textwidth}{!}{
\begin{tabular}{l||cccc|l|l|l}
\toprule
\rowcolor{lightgray}
    \rowcolor{lightgray}
    \multicolumn{1}{l||}{\multirow{-1}{*}{Methods}} 
    & \multicolumn{1}{c}{\multirow{-1}{*}{Stride}}
    & \multicolumn{1}{c}{\multirow{-1}{*}{Avgpool}}
    & \multicolumn{1}{c}{\multirow{-1}{*}{Bicubic}}
    & \multicolumn{1}{c|}{\multirow{-1}{*}{WFD}}
    & \multicolumn{1}{c|}{\multirow{-1}{*}{Params}}
    & \multicolumn{1}{c|}{\multirow{-1}{*}{FLOPs}}
    & \multicolumn{1}{c}{\multirow{-1}{*}{PSNR / SSIM}}

    \\ 
    \hline\hline
    Case1     & \Checkmark   & \textcolor{gray}{\XSolidBrush}  & \textcolor{gray}{\XSolidBrush} & \textcolor{gray}{\XSolidBrush}  & 0.830M & 1.131G & 26.22 / 0.7743 \\ 
    Case2     & \textcolor{gray}{\XSolidBrush}   & \Checkmark  & \textcolor{gray}{\XSolidBrush} & \textcolor{gray}{\XSolidBrush}  & 0.830M & 1.129G & 26.26 / 0.7747 \\ 
    Case3     & \textcolor{gray}{\XSolidBrush}   & \textcolor{gray}{\XSolidBrush}  & \Checkmark & \textcolor{gray}{\XSolidBrush}  & 0.830M & 1.129G & 26.21 / 0.7731 \\ 
    \textbf{Ours}     & \textcolor{gray}{\XSolidBrush}   & \textcolor{gray}{\XSolidBrush}  & \textcolor{gray}{\XSolidBrush} & \Checkmark  & 0.848M & 1.164G & \bf{26.36 / 0.7795} \\ 

    \bottomrule
\end{tabular}}
\end{table}
\hspace{-2mm}

\section{More Ablation Studies}
\paragraph{Downsampling Chosen}
In the main text, we only provide an ablation study using stride convolution instead of our proposed wavelet feature downsample (WFD) module for downsampling. TABLE~\ref{tab:supp_ablation} is additionally supplemented with experimental results of encoder stage downsample using average pool (Avgpool), and bicubic downsample (Bicubic). Compared with these common downsample strategies, our proposed WFD module can minimize the adverse effects of downsample on FSR reconstruction, achieving a performance improvement of more than 0.1dB PSNR with only a small number of params and FLOPs gains. This experiment demonstrates the advantages of our proposed WFD over existing downsample strategies. Additionally, it also confirms that reducing the feature corruption caused by downsampling in the encoding stage can significantly improve the model's performance and further enhance the model's efficiency.

\paragraph{Generalization of Our Method}
We randomly select 50 face images with diverse lights and occlusions from unused parts of the CelebA~\cite{liu2015deep} dataset. TABLE~\ref{tab:generalizability} and Fig.~\ref{fig:review1_scenary} show comparisons of quantitative and qualitative FSR results, respectively, where ours are superior to existing methods. 
\paragraph{Results of Face Alignment}
As for face alignment, we use a face alignment library~\cite{bulat2017far} to evaluate coordinate errors of 68 facial landmarks on reconstructed faces from the Helen [13] test set. TABLE~\ref{tab:face_alignment} shows that our method reconstructs face images with minimized average coordinate error.
\paragraph{Visual Results for Full-domain Cases}
We provide a visualization of our full-domain cases in Fig.~\ref{fig:review2_full_domain} to complement the persuasiveness of ablations. Our method can recover accurate faces due to the integration of local, regional, and global facial features.

\begin{table}[t!]
\tiny
\setlength\tabcolsep{3.5pt}
\centering
\vspace{0mm}
    \caption{Quantatitive evaluation of model generalizability.}
\vspace{-1mm}
\label{tab:generalizability}
\resizebox{0.48\textwidth}{!}{
\begin{tabular}{l||cccc}
\toprule
\rowcolor{lightgray}
& \multicolumn{4}{c}{PSNR$\uparrow$/SSIM$\uparrow$} 
\\ 
\cmidrule{2-5}
    \rowcolor{lightgray}
    \multicolumn{1}{l||}{\multirow{-2}{*}{Scenario}} 
    & SPARNet~\cite{chen2020learning}  & AD-GNN~\cite{bao2022attention}  & SFMNet~\cite{wang2023spatial} & \textbf{Ours}   
    \\ 
    \hline\hline
    Diverse-lights     & 29.34/0.8160   & 29.67/0.8214  & 29.72/0.8232  & \textbf{29.86/0.8275}         \\ 
    Occlusions     & 26.48/0.7082  & 26.59/0.7096  & 26.69/0.7138  & \textbf{26.79/0.7210}        \\ 
    \bottomrule
\end{tabular}}
\hspace{-1mm}
\end{table}

\begin{table}[t!]
\tiny
\setlength\tabcolsep{2.5pt}
\centering
\vspace{-1mm}
    \caption{Quantatitive evaluation of face alignment.}
\vspace{-1mm}
\label{tab:face_alignment}
\resizebox{0.48\textwidth}{!}{
\begin{tabular}{l||ccccc}
\toprule
\rowcolor{lightgray}
    \multicolumn{1}{l||}{\multirow{-1}{*}{Average Error}} 
    & LR & SPARNet~\cite{chen2020learning}  & AD-GNN~\cite{bao2022attention}  & SFMNet~\cite{wang2023spatial} & \textbf{Ours}   
    \\ 
    \hline\hline
    $\Delta$(x, y)$\downarrow$  & (7.41, 13.31)   & (1.25, 1.75)   & (1.33, 1.72)  & (1.18, 1.56)  & \textbf{(1.08, 1.36)}         \\ 
    \bottomrule
\end{tabular}}
\hspace{-1mm}
\end{table}

\begin{figure}[t!]
\begin{center}
\vspace{-4mm}
\includegraphics[width=0.97\linewidth]{rebuttal/review1_scenary.pdf}
\vspace{-1mm}
\end{center}
\caption{Qualitative evaluation of model generalizability.}
\label{fig:review1_scenary}
\vspace{-1mm}
\end{figure}

\begin{figure}[t!]
\begin{center}
\includegraphics[width=0.97\linewidth]{rebuttal/review2_full_domain.pdf}
\vspace{-1mm}
\end{center}
\caption{Qualitative evaluation of full-domain cases.}
\label{fig:review2_full_domain}
\vspace{-1mm}
\end{figure}

\section{Discussion.} Experimental results show that the efficiency of our method is contributed by two parts: wavelet-based encoder-decoder structure and full-domain Transformer. Both parts can be integrated into existing encoder-decoder-based methods~\cite{chen2020learning,wang2023spatial,LAATransformer} to further enhance their performance. Therefore, our method is not only efficient in performance but also generalizable.

\section{Limitation}
Our method aims to preserve identity accuracy. It can restore face images with higher fidelity compared to existing generative prior-based methods, but its clarity is not as sharp as theirs. In this context, we will further discuss how to integrate the generative prior based on this method to enhance clarity while maintaining the identity accuracy of the restored face images.

\bibliographystyle{ACM-Reference-Format}
\balance
\bibliography{sample-base}

\end{sloppypar}